\pdfoutput=1
\documentclass{article}

\usepackage[final]{neurips_2024}

\usepackage{times}
\usepackage{epsfig}
\usepackage{graphicx}
\usepackage{float}
\usepackage{wrapfig}
\usepackage{amsmath,amssymb,amsthm}
\usepackage{algorithm,algorithmicx,algpseudocode}
\usepackage{bm,xspace}
\usepackage{comment}
\usepackage{verbatim}
\usepackage{multirow}
\usepackage{balance}
\usepackage{url}
\usepackage{booktabs}
\usepackage{etoolbox,siunitx}
\usepackage{calc}
\usepackage{pifont,hologo}
\usepackage{color}
\usepackage{ulem}

\newcolumntype{P}[1]{>{\centering\arraybackslash}p{#1}}
\newcolumntype{M}[1]{>{\centering\arraybackslash}m{#1}}

\usepackage[super]{nth}
\usepackage{nicefrac}
\robustify\bfseries

\DeclareMathSymbol{@}{\mathord}{letters}{"3B}

\def\latex/{\LaTeX}
\def\bibtex/{\hologo{BibTeX}}

\makeatletter
\DeclareRobustCommand\onedot{\futurelet\@let@token\@onedot}
\def\@onedot{\ifx\@let@token.\else.\null\fi\xspace}

\newcommand*{\Rom}[1]{\expandafter\@slowromancap\romannumeral #1@}
\newcommand*{\rom}[1]{\expandafter\romannumeral #1}

\def\eqref#1{equation~\ref{#1}}

\def\1{\bm{1}}

\DeclareMathAlphabet{\mathsfit}{\encodingdefault}{\sfdefault}{m}{sl}
\SetMathAlphabet{\mathsfit}{bold}{\encodingdefault}{\sfdefault}{bx}{n}

\usepackage[table]{xcolor}
\definecolor{blue}{HTML}{0055cc}
\definecolor{red}{HTML}{cc1100}
\definecolor{orange}{HTML}{cc7700}
\definecolor{green}{HTML}{339955}
\definecolor{Highlight}{rgb}{0.12,0.49,0.85}
\usepackage{float}
\usepackage[utf8]{inputenc} 
\usepackage[T1]{fontenc}    
\usepackage[pagebackref,breaklinks,colorlinks,linkcolor=purple,citecolor=Highlight]{hyperref}       
\usepackage{url}            
\usepackage{booktabs}       
\usepackage{amsfonts}       
\usepackage{nicefrac}       
\usepackage{microtype}      
\usepackage{xcolor}         
\usepackage{amsmath}        
\usepackage{amssymb}        
\usepackage{lipsum}         
\usepackage{bbding}         
\usepackage{array}          
\usepackage{subcaption}     
\usepackage{pdfpages}
\usepackage{natbib}
\usepackage{ulem}
\usepackage{makecell}

\usepackage{caption}
\setcitestyle{square,numbers,comma,sort}
\usepackage[capitalize]{cleveref}
\crefname{section}{Sec.}{Secs.}
\Crefname{section}{Section}{Sections}
\Crefname{table}{Table}{Tables}
\crefname{table}{Tab.}{Tabs.}

\newlength\savewidth

\renewcommand{\paragraph}[1]{\noindent\textbf{#1}}

\newcolumntype{x}[1]{>{\centering\arraybackslash}p{#1pt}}
\newcolumntype{y}[1]{>{\raggedright\arraybackslash}p{#1pt}}
\newcolumntype{z}[1]{>{\raggedleft\arraybackslash}p{#1pt}}

\newcommand{\app}{\raise.17ex\hbox{$\scriptstyle\sim$}}

\definecolor{deemph}{gray}{0.6}

\definecolor{baselinecolor}{gray}{.9}

\definecolor{my_red}{HTML}{FE4444}
\definecolor{Gray}{gray}{0.95}

\newcommand{\improve}[1]{{\textbf{\textcolor{green}{{#1}}}}}

\newcommand{\improvetit}[1]{{\textit{\textbf{\textcolor{green}{{#1}}}}}}

\let\originalleft\left
\let\originalright\right
\renewcommand{\left}{\mathopen{}\mathclose\bgroup\originalleft}
\renewcommand{\right}{\aftergroup\egroup\originalright}

\title{ReFIR: Grounding Large Restoration Models \\ with Retrieval Augmentation}

\author{%
   Hang Guo$^{1}$ \quad  
   Tao Dai\footnotemark[1] $^{\ 2}$ \quad 
   Zhihao Ouyang$^{3}$ \quad 
    Taolin Zhang$^{1}$ \\ 
    \textbf{ Yaohua Zha$^{1}$ \quad 
    Bin Chen$^{4}$ \quad 
    Shu-tao Xia$^{1,5}$ } \vspace{0.1cm} \\
    $^1$Tsinghua University \quad
    $^2$Shenzhen University \quad
    $^3$Aitist.ai \\ 
    $^4$Harbin Institute of Technology \quad 
    $^5$Peng Cheng Laboratory  \vspace{0.1cm} \\ 
  \url{https://github.com/csguoh/ReFIR}
}

\begin{document}
\renewcommand{\thefootnote}{\fnsymbol{footnote}}
\footnotetext[1]{Corresponding author: Tao Dai}

\maketitle

\vspace{-3mm}
\begin{abstract}

\vspace{-2mm}
Recent advances in diffusion-based Large Restoration Models (LRMs) have significantly improved photo-realistic image restoration by leveraging the \textit{internal knowledge} embedded within model weights. However, existing LRMs often suffer from the \textit{hallucination} dilemma, $\textit{i.e.}$, producing incorrect contents or textures when dealing with severe degradations, due to their heavy reliance on limited internal knowledge. In this paper, we propose an orthogonal solution called the \textbf{Re}trieval-augmented \textbf{F}ramework for \textbf{I}mage \textbf{R}estoration (ReFIR), which incorporates retrieved images as \textit{external knowledge} to extend the knowledge boundary of existing LRMs in generating details faithful to the original scene.  Specifically, we first introduce the nearest neighbor lookup to retrieve content-relevant high-quality images as reference, after which we propose the cross-image injection to modify existing LRMs to utilize high-quality textures from retrieved images. Thanks to the additional external knowledge, our ReFIR can well handle the hallucination challenge and facilitate faithfully results. Extensive experiments demonstrate that ReFIR can achieve not only high-fidelity but also realistic restoration results. Importantly, our ReFIR requires no training and is adaptable to various LRMs.

\end{abstract}

\vspace{-3mm}


\section{Introduction}
\vspace{-2mm}

Restoring a high-quality image (HQ) from its low-quality counterpart (LQ) is a well-known ill-posed problem and has been studied over the years~\cite{dong2014learning,dai2019second,guo2024mambair,zhang2017learning,chen2021pre,li2021efficient,liang2021swinir,wang2021real,zhang2021designing}. Previous efforts attempt to handle this problem through employing various neural network architectures, including CNNs, GANs and Transformers. Recently, diffusion models~\cite{ho2020ddpm,song2020ddim} have emerged as a promising alternative, delivering noteworthy results in real-world image restoration~\cite{fei2023generative,kawar2022denoising,wang2022zero}. In particular, some works~\cite{lin2023diffbir,wang2023stablesr,yang2023pasd,sun2023ccsr,wu2023seesr,yu2024supir} have successfully leveraged the powerful generative prior of pre-trained text-to-image (T2I) diffusion models for scaling up, to obtain the Large Restoration Model (LRM) with billions of parameters, bringing significant progress in restoring photo-realistic images.

\begin{figure}[!t]
\centering
    \includegraphics[width=0.99\linewidth]{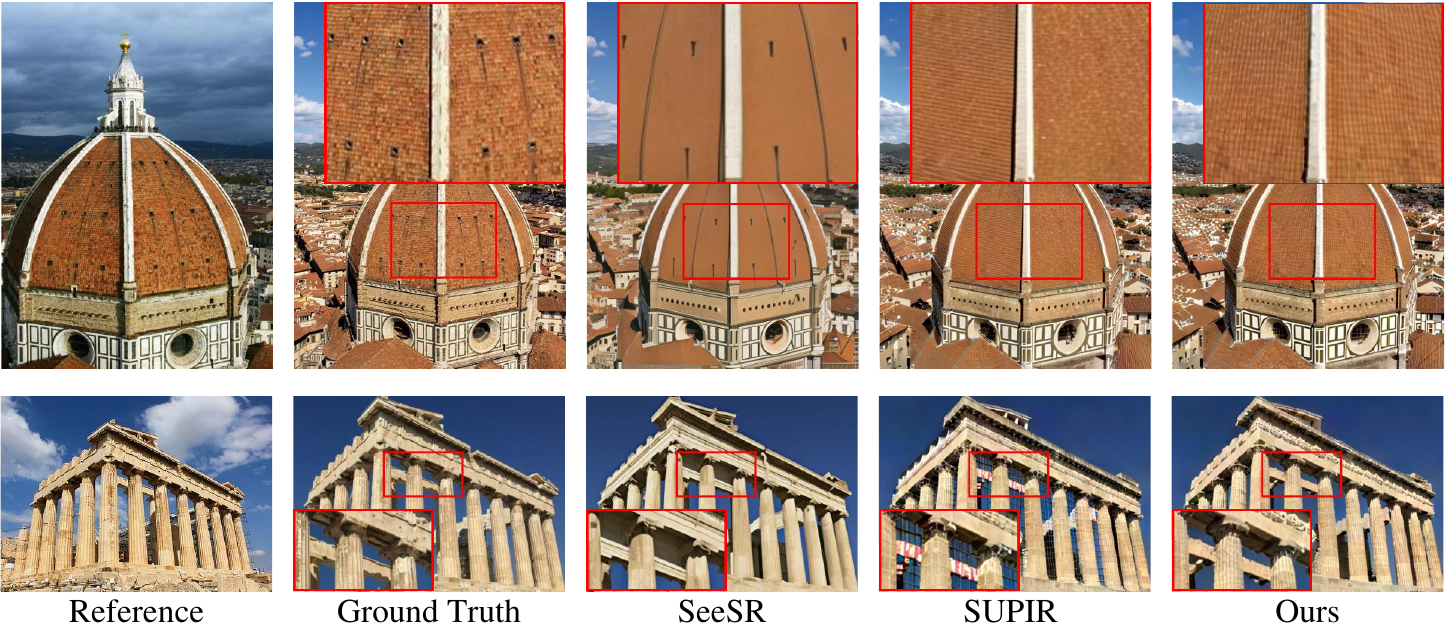}
    \vspace{-2mm}
    \caption{Existing LRMs encounter hallucination issues,  \textit{i.e.}, generating contents or details that deviate from the original scene, when dealing with challenging degradations. By incorporating the proposed ReFIR to existing LRMs~\cite{yu2024supir} without any training, the additional external knowledge facilitates producing more faithful results. Please zoom in for better visualization.}
    \label{fig:motivation}
    \vspace{-12mm}
\end{figure}

Although scaling up restoration models has achieved remarkable success, existing LRMs may not always produce results that are faithful to the original scene, particularly when faced with heavily degraded images that surpass the LRMs' capabilities (see \cref{fig:motivation}). This issue is similar to the hallucination problem observed in large language models (LLMs)~\cite{min2023factscore,mallen2022not}, \textit{e.g.} ChatGPT might generate nonsense responses when highly specialized questions exceed its knowledge boundary. Similarly, if one LRM has never seen a specific scene, it will struggle to restore corresponding images faithfully. By analogizing LLM to LRM, we define the phenomenon where LRMs generate textures inconsistent with the original scene when facing hard samples as the hallucination of LRMs.

To address the hallucination problem in LRMs, simply expanding the \textit{internal knowledge} through additional training data and parameters might seem straightforward, but it can significantly increase computational and storage costs. Instead, this work considers another orthogonal strategy that enhances the \textit{external knowledge} of LRMs without adding parameter counts. Drawing inspiration from the retrieval-augmented generation (RAG) used in LLMs~\cite{ram2023context,asai2023self,asai2023retrieval}, we aim to use the retrieved high-quality content-relevant images as external knowledge to alleviate the hallucination of LRMs. However, applying RAG to image restoration poses specific challenges. Specifically, in natural language, simply feeding the retrieved documents along with the original user query to LLMs can allow it to produce grounded responses. However, in the context of image restoration, allowing low-quality images to attend to retrieved images during their restoration process is non-trivial, which motivates us to develop novel techniques to enable LRMs to utilize external knowledge in restoration.

To this end, we delve deep into the working mechanisms of LRMs for insightful observations. Details of the experimental setup are described in \cref{sec:observation}. Our key findings indicate that the workflow of LRMs can be divided into two distinct stages: the \textbf{Denoising Structure Reconstruction} stage, during which the self-attention in the ControlNet~\cite{zhang2023adding} reconstructs a clear overall structure from the noised representation. After that, in the \textbf{Detail Texture Restoration} stage, the self-attention in the UNet~\cite{ronneberger2015unet} decoder fills scene-specific textures based on the denoised structure map. Based on these findings, a natural solution emerged: we can transfer high-quality, scene-specific textures from the retrieved images to the low-quality images during the detail texture restoration stage. In this way, the restored image is allowed a consistent texture with the retrieved image, thus mitigating the hallucination.

Inspired by the above observation, in this work, we propose the Retrieval-augmented Framework for Image Restoration, dubbed ReFIR, to offer a simple but effective way to expand the knowledge boundary of LRMs using the external knowledge from the retrieved images. Specifically, we first construct the retriever which employs the nearest neighbor lookup in the semantic embedding space to retrieve content-relevant reference images in the high-quality image database. After that, we develop the cross image injection which modifies the self-attention layer of original LRMs to enable the queries from the low-quality denoising chain to attend to the keys and values from the denoising chain of retrieved reference. To avoid the domain preference problem during injection, we propose separate attention to perform intra-chain and inter-chain attention, respectively. Given the spatial misalignment between the LQ and the retrieved HQ, we further adopt spatial adaptive gating to mask meaningless pixels during injection. At last, we employ the distribution alignment to narrow the domain gap between LQ and retrieved images. Thanks to the proposed ReFIR, the restoration of the LQ image can make full use of the external knowledge from the reference to generate high-fidelity images. Notably, the proposed pipeline is training-free and can be applied to multiple LRMs.

The contribution of this paper can be summarized as: 
\textbf{(i)} We introduce retrieval-augmented restoration, a novel concept to mitigate the hallucination problems in existing LRMs.
\textbf{(ii)} We conduct an in-depth analysis of the working mechanisms of LRMs,  based on which we propose a training-free framework to utilize the retrieved images.
\textbf{(iii)} Extensive experiments validate that our proposed method effectively mitigates hallucination and is applicable to a broad spectrum of existing LRMs.


\section{Related Works}

\subsection{Diffusion Model for Image Restoration}
Diffusion models have recently achieved significant advancements across various computer vision tasks~\cite{peebles2023scalable,cao2023masactrl,zhang2024real,gu2024photoswap,blattmann2022retrieval}. In the realm of image restoration, early explorations often involved training diffusion models from scratch to obtain the restoration tailored models~\cite{rombach2022high,shang2024resdiff, yue2024resshift, zhang2024unified}. While these models are capable of producing high-fidelity results, they usually fall short of generating perceptually pleasing images. To leverage the powerful generative capabilities of large pre-trained text-to-image diffusion models like Stable Diffusion~\cite{rombach2022high}, recent attempts~\cite{lin2023diffbir,wang2023stablesr,yang2023pasd,sun2023ccsr,wu2023seesr,yu2024supir} have focused on using the ControlNet~\cite{zhang2023controlnet} with a LQ image as the condition to generate HQ images. Benefiting from the scaling law~\cite{kaplan2020scaling}, these large restoration models with billions of parameters have shown impressive restoration results with photo-realistic textures and details. However, similar to the large language models, when the user query, \textit{i.e.}, the LQ image in this setting, exceeds the knowledge boundary of the large models, the models often fail to generate meaningful or correct responses, which is unacceptable for image restoration tasks that pursue high-fidelity.

\begin{figure}[!t]
    \centering
    \includegraphics[width=1.\linewidth]{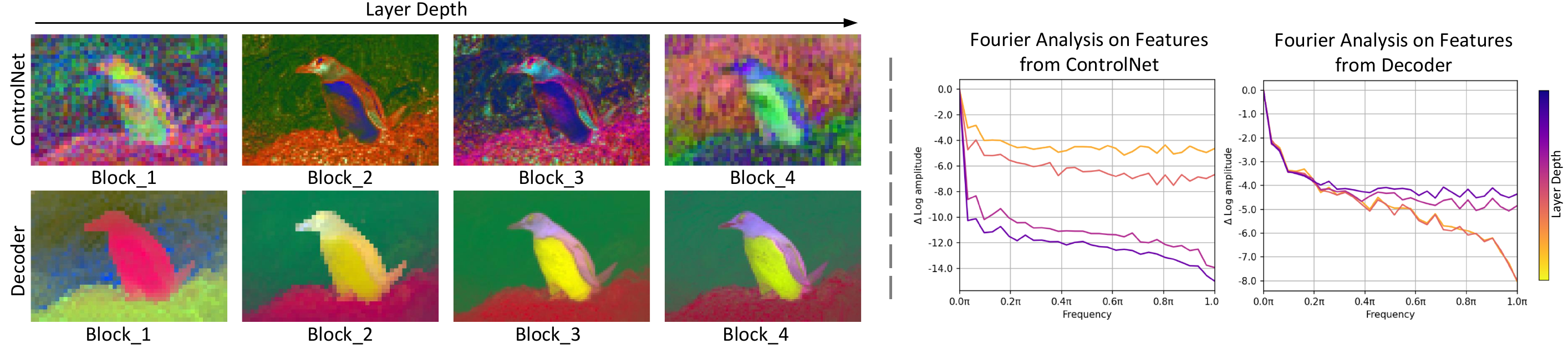}
    \caption{In-depth visualization about the working mechanism of LRM. \textbf{Left}: we use PCA to visualize the top three principal components of latent extracted from the self-attention layer of the ControlNet and UNet decoder. 
    \textbf{Right}: quantitative power spectrum of the corresponding latent using Fourier analysis. More visualization can be found in~\cref{sec:supp-additional-viz-res}.
    }
    \vspace{-5mm}
    \label{fig:observation}
\end{figure}

\subsection{Reference-based Image Super-resolution}

Compared with single image super-resolution~\cite{dong2014learning,zhang2017learning,dai2019second}, Reference-based Image Super-Resolution (RefSR) can achieve enhanced performance by employing content-similar reference images as the additional input, and has attracted great research interests in the past few years~\cite{yang2020learning,lu2021masa,jiang2021robust}. For instance, C2-Matching~\cite{jiang2021robust} introduces a teacher-student correlation distillation and a dynamic DCN aggregation module for more precise alignment between low-quality and reference images. Following this, DATSR~\cite{cao2022reference} employs reciprocal learning and SwinTransformer to further boost performance. Additionally, MRefSR~\cite{zhang2023lmr} introduces a simple baseline to facilitate RefSR with multiple reference images. It is worth mentioning that despite both using additional images as references, our proposed retrieval augmented restoration pipeline differs from previous RefSR methods in several key aspects. Firstly, current RefSR models are typically small-scale due to limited training data, leading to performance degradation under challenging real-world conditions. Secondly, most RefSR methods can only use one single reference image and even fail to work in the absence of reference images. Thirdly, different from RefSR models that require training, our method can inject image-specific external knowledge into LRMs in a training-free manner. We give a detailed discussion about the difference in~\cref{sec:supp-difference-RefSR}.

\subsection{Retrieval Augmented Generation}

In the domain of natural language processing, Retrieval-Augmented Generation (RAG) leverages the strengths of pre-trained Large Language Models (LLMs) combined with knowledge retrieved from an external document database to enhance the quality of generated content~\cite{min2023factscore,mallen2022not}. Typically, a RAG system initially retrieves documents relevant to the user's query from the knowledge base and then integrates the retrieved document along with the original user query into the LLMs without any tuning to generate a response. Even when no relevant document is available, this system can still operate by using the internal knowledge embedded in the LLMs' parameters. The integration of RAG allows LLMs to produce outputs that are not only contextually rich but also factually accurate, effectively mitigating the hallucination problem in knowledge-intensive tasks~\cite{ram2023context,asai2023self,asai2023retrieval}. In this work, we extend the concept of RAG to image processing and propose retrieval-augmented restoration to alleviate the hallucination issues in LRMs. By utilizing external textures embedded in the retrieved reference images, our tuning-free framework significantly facilitates faithful restoration results.

\begin{figure}[!t]
    \centering
    \includegraphics[width=1.\linewidth]{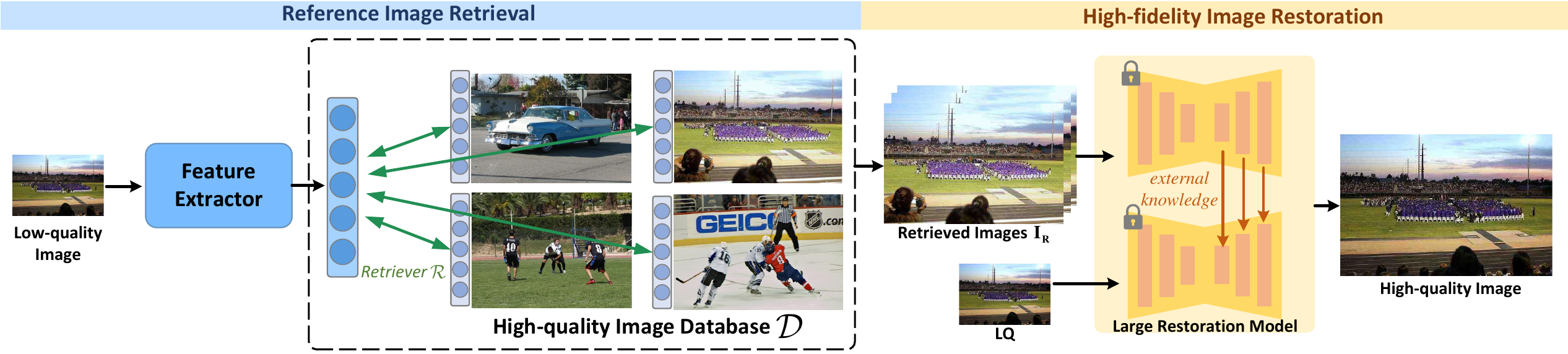}
    \caption{Our ReFIR consists of two stages: the \textbf{Reference Image Retrieval} stage employs the retriever $\mathcal{R}$ to search content-relevant images from high-quality image database $\mathcal{D}$, and then the \textbf{High-fidelity Image Restoration} stage restores HQ image with reference images $\mathbf{I_R}$ as condition. The proposed framework is highly generic and can be applied to multiple existing LRMs without any training or fine-tuning. }
    \vspace{-5mm}
    \label{fig:pipeline}
\end{figure}


\section{Probing Large Restoration Models}
\label{sec:observation}
In order to manipulate the LRM so that it can utilize the retrieved reference images as external knowledge, we first delve into the underlying mechanism of existing LRMs to find useful insights. We choose the current popular LRM method SUPIR~\cite{yu2024supir} as a representative. Inspired by previous image editing efforts ~\cite{hertz2022prompt,gu2024photoswap,zhang2024real,cao2023masactrl}, which show that the self-attention layer of diffusion models contains important spatial correlation of an image, we thus follow this clue and employ the PCA to visualize the principal components of the latent from self-attention layers of SUPIR. We further utilize the Fourier analysis~\cite{park2021how} to allow for quantitative results. The results are shown in \cref{fig:observation}.

It can be seen that the ControlNet of the LRM can denoise the latent as the layers deepen, facilitating the reconstruction of a clear overall structure. However, this process is accompanied by a reduction in the high-frequency meaningful texture of the original image. This qualitative visualization can be also verified by the frequency characteristic plots, with high-frequency components decaying as layer number increases. 
On the other hand, the role of the UNet decoder is significantly different. Based on the previous clear structural map, the decoder restores the high-frequency details and textures with the help of skip connections, which is also shown through the strengthening high-frequency component in the decoder's frequency curve.

Considering the above observations, we can divide the image restoration process of the LRM into two phases: the Denosing Structure Reconstruction phase in the ControlNet, and the Detail Texture Restoration phase in the UNet decoder. Inspired by these probing experiments, in this work, we employ the detail texture restoration nature in the self-attention layer of the decoder to inject the high-fidelity textures of retrieved images into the restoration process of the low-quality image.

\section{Methodology}

This work considers using retrieved reference images as an explicit part of the model. In contrast to the existing restoration pipeline, our ReFIR is parameterized by not only the internal knowledge from the network weights but also the external knowledge retrieved from suitable data representations. \cref{fig:pipeline} gives an overview of our ReFIR. In the following part, we will first give the technical details of the retriever for reference image retrieval in \cref{sec:retrive-pipeline}, followed by the cross image injection to inject the external data knowledge into the restoration process of LRMs in \cref{sec:cross-injection}.

\begin{figure}[!t]
    \centering
    \includegraphics[width=1.\linewidth]{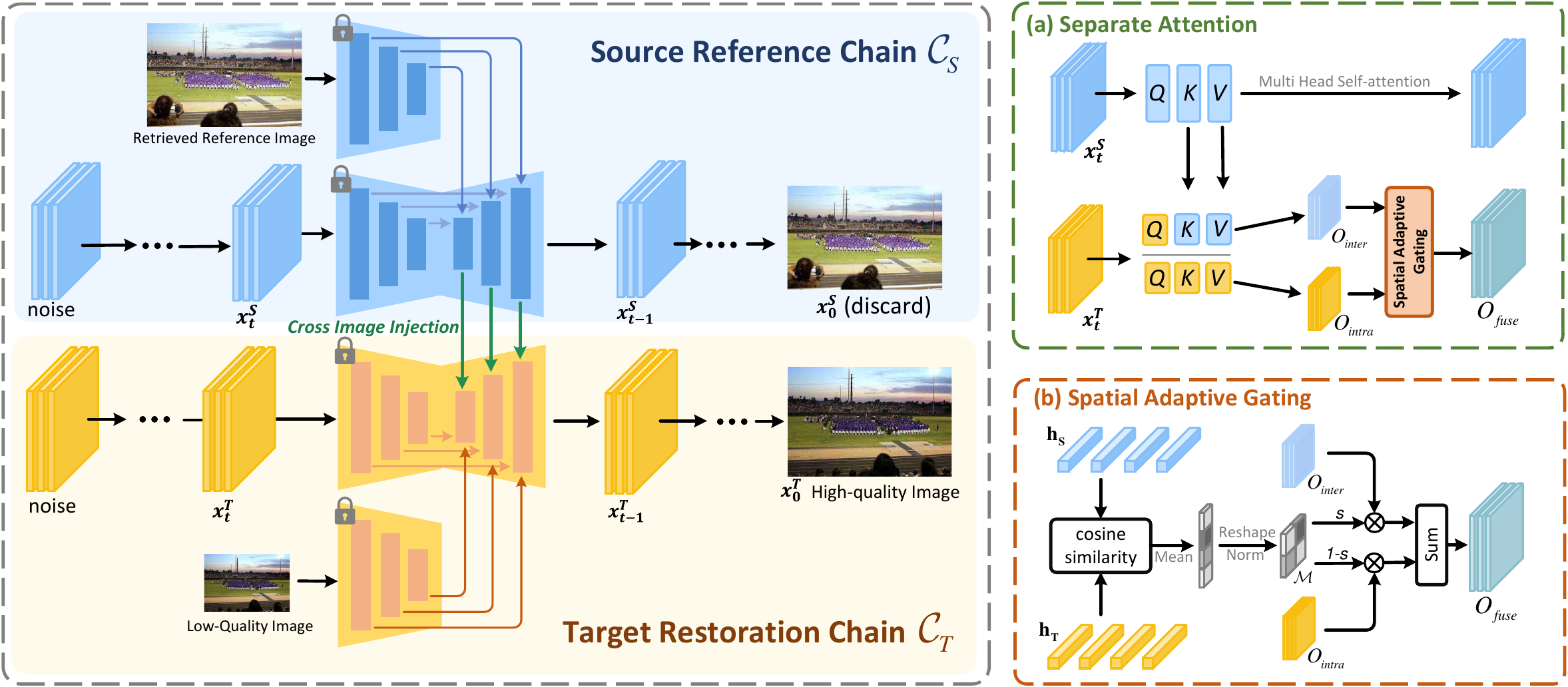}
    \caption{An illustration of cross image injection. Both $\mathcal{C}_T$ and $\mathcal{C}_S$ share the same model weights.}
    \vspace{-3mm}
    \label{fig:cross-img-injection}
\end{figure}

\subsection{Nearest Neighbor Lookup for Reference Image Retrieval}
\label{sec:retrive-pipeline}

Our reference image retrieval system can be represented as a binary set $\{\mathcal{D},\mathcal{R}\}$, where $\mathcal{D}$ is a fixed database containing a large number of HQ images, and $\mathcal{R}$ denotes a non-parametric retriever to obtain the retrieved image set $\mathbf{I_R}$ which consists of $k$ elements and is a subset of $\mathcal{D}$ given a query LQ image $I_{LQ} \in \mathbb{R}^{3\times H \times W}$, \textit{i.e.}, $\mathcal{R}: I_{LQ}, \mathcal{D} \mapsto \mathbf{I_R}$, where $\mathbf{I_R} \subseteq \mathcal{D}$ and $|\mathbf{I_R}|=k$. Ideally, $\mathcal{R}$ has to be designed such that it provides the model with beneficial data representations from $\mathcal{D}$ to help restore images containing details faithful to the original scenes.

In this work, we implement a conceptually simple solution of $\mathcal{R}$, which uses the query image $I_{LQ}$ to retrieve its $k$ nearest neighbor in $\mathcal{D}$ using cosine similarity in the compact feature space derived from any feature extractors, such as VGG~\cite{simonyan2014very}, ResNet~\cite{he2016deep} or CLIP~\cite{radford2021learning}. Since the $\mathcal{D}$ is fixed, in practice, we can pre-extract and store the compact feature before training. Given a sufficiently large database $\mathcal{D}$, this strategy ensures that the set of neighbors $\mathbf{I_R}$ shares sufficient semantic consistency with $I_{LQ}$ and thus provides useful visual information for the restoration. Although this scheme seems simple, we show that it is efficient and effective, please see \cref{sec:supp-effctive-retriver} for discussion.

\subsection{Cross Image Injection for High-fidelity Image Restoration}
\label{sec:cross-injection}
Given the retrieved reference images $\mathbf{I_R}=\mathcal{R}(I_{LQ},\mathcal{D})$, we further propose the cross image injection to allow the original LRMs to use the external knowledge from $\mathbf{I_R}$. 
As shown in \cref{fig:cross-img-injection}, we first construct two parallel denoising chains: the target restoration chain $\mathcal{C}_{T}$ which is used to restore $I_{LQ}$, and the source reference chain $\mathcal{C}_S$ which unfolds $\mathbf{I_{R}}$ into denoising time steps. After that, we introduce separate attention to separately perform attention within and between chains, followed by spatial adaptive gating to filter out irrelevant pixels. At last, we use the distribution alignment to mitigate the domain gap between chains. More details are given below.






\noindent
\textbf{Separate attention.}
To allow the $\mathcal{C}_T$ to learn the knowledge from the $\mathcal{C}_S$, an effective interaction between the latents is crucial. 
Inspired by the observation in \cref{sec:observation}, we aim to transfer the knowledge embedded in the self-attention layer of $\mathcal{C}_S$'s decoder to the counterpart of $\mathcal{C}_T$.
%
To this end, we modify the original self-attention in $\mathcal{C}_T$ to our proposed separate attention. 
The core idea of our separate attention is to add ``inter-chain cross-attention” to the original ``intra-chain self-attention” so that $\mathcal{C}_T$ can attend high-quality texture knowledge from $\mathcal{C}_S$ while preserving its original features.
As shown in~\cref{fig:cross-img-injection}(a), formally, denote $Q_T$, $K_T$, $V_T$ as the query, key and value from the $\mathcal{C}_T$, and $K_S$, $V_S$ as the key and value from the $\mathcal{C}_S$, the intra-chain self-attention preserves the original attention of $\mathcal{C}_T$ to obtain the output $O_{intra}$, and the inter-chain cross-attention uses the $Q_T$ to query the $K_S$ and $V_S$ to facilitate $\mathcal{C}_T$ utilizing the knowledge from $\mathcal{C}_S$ to get the result $O_{inter}$. In short, the proposed separate attention can be formalized as follows:
\begin{equation}\begin{aligned}
\label{eq:decoupled-attn}
O_{intra} =  \mathtt{Attention}(\textcolor[RGB]{204, 106, 28}{Q_T}, \textcolor[RGB]{204, 106, 28}{K_T}, \textcolor[RGB]{204, 106, 28}{V_T}),\quad O_{inter} = \mathtt{Attention}(\textcolor[RGB]{204, 106, 28}{Q_T}, \textcolor[RGB]{88,155,222}{K_S},\textcolor[RGB]{88,155,222}{V_S}).
\end{aligned}\end{equation}
\noindent

It is worth mentioning that directly using $Q_T$ to query the concatenate results of $K_T$ and $K_S$ 
can only yield sub-optimal results due to the domain preference issue, \textit{i.e.}, $Q_T$ will prefer latent from the same domain $\mathcal{C}_T$ even though $\mathcal{C}_S$ is more helpful for reconstruction. By using the proposed separate attention, the $Q_T$ is separated to attend $K_T$ and $K_S$, thus effectively mitigating this problem. We give more discussion in \cref{sec:diss-separate-attention}.

\noindent
\textbf{Spatial adaptive gating.}
We then consider fusing the separate attention results $O_{intra}$ and $O_{inter}$. The main challenge is the spatial misalignment between $I_{LQ}$ and $\mathbf{I_R}$. For instance, the same objects may appear in different locations in $I_{LQ}$ and $\mathbf{I_R}$, or some objects in $I_{LQ}$ may not present in $\mathbf{I_{R}}$ and vice versa. As a result, some pixels in $Q_T$ may not find the corresponding reference in $K_S$, resulting in some pixels in $O_{inter}$ meaningless.

To address this spatial misalignment, we propose the spatial adaptive gating to selectively fuse $O_{intra}$ and $O_{inter}$ without introducing additional parameters (\cref{fig:cross-img-injection}(b)). Specifically, given latents at specific denoising blocks from $\mathcal{C}_T$ and $\mathcal{C}_S$, respectively, we first flatten them along the spatial dimension to obtain $\mathbf{h_T},\mathbf{h_S} \in \mathbb{R}^{C \times HW}$. Next, we compute their pixel-wise cosine similarity to obtain the similarity matrix $\mathrm{sim} \in \mathbb{R}^{HW \times HW}$. Since the $i$-th row of $\mathrm{sim}$ represents the similarity of the $i$-th pixel in $\mathbf{h_T}$ to all the pixels in $\mathbf{h_S}$, therefore, a large sum of the $i$-th row indicates a large impact of $\mathbf{h_S}$ in restoring the $i$-th pixel of $\mathbf{h_T}$. Following this idea, we summation over the $i$-th row of the $\mathrm{sim}$ to approximate the utility of $\mathbf{h_S}$ to the $i$-th pixel of $\mathbf{h_T}$. Finally, we reshape this summation results back to 2D shape and use min-max normalization to restrict the range to $[0, 1]$, to get the pixel-wise mask $\mathcal{M}$ for adaptive gated fusion: 
\begin{equation}
\label{eq:scale-s}
O_{fuse} = (\mathbf{1}-s \mathcal{M})\otimes O_{intra} + s \mathcal{M} \otimes O_{inter},
\end{equation}
\noindent 
where $s$ is a user-defined scalar to control the degree to which the restored image attends the retrieved images, $\otimes$ denotes the Hardamard product, and $\mathbf{1}$ is an all one tensor with the same shape as $\mathcal{M}$.

\noindent \textbf{Distribution alignment.}
Using the $O_{fuse}$ to replace the original intra-chain self-attention results $O_{intra}$ seems to be a promising way to integrate useful external knowledge from $\mathcal{C}_S$.
However, it should be noticed that there is a domain gap between $\mathcal{C}_T$ and $\mathcal{C}_S$ due to the image quality and content differences, and thus 
a direct insertion of $O_{intra}$ into $\mathcal{C}_T$ will result in a distribution shift of the original denoising chain in $\mathcal{C}_T$.

To this end, we propose the distribution alignment as a complementary to calibrate the distribution shift. Specifically, considering the latent in the diffusion chain is a Gaussian, we propose to use the Adaptive Instance Normalization (AdaIN)~\cite{huang2017adain} to align the mean and variance of  $O_{fuse}$ to the original statistics of $O_{intra}$: 
\begin{equation}
O'_{fuse} = \mathtt{AdaIN}(O_{fuse}, O_{intra}),
\end{equation}
\noindent
where $ \mathtt{AdaIN}(u,v)$ denotes replacing the mean and variance of $u$ with the corresponding part of $v$. Finally, we replace the original self-attention result in $\mathcal{C}_T$ with the well-aligned $O'_{fuse}$ to finish the cross image injection process.

\section{Experiments}

\subsection{Experiments Setup}

\noindent \textbf{Datasets and metrics.} 
In this work, we include experiments with two difficulty levels for performance evaluation. The first setup considers restoration with manually provided ideal reference images, which share a high content similarity with the LQ image, to evaluate the ability to utilize the reference knowledge. The datasets for this setting employ the widely used RefSR dataset including CUFED5~\cite{zhang2019imagecufed,wang2016event} and  WR-SR~\cite{jiang2021robust}, in which the reference images are already provided. Since these datasets only contain HQ images, we thus use the second-order degradation model from Real-ESRGAN~\cite{wang2021real} with $\times 4$ down-sampling scale to generate the real-world degraded images. The second setup turns to more challenging practice where the reference images have to be retrieved using the retriever, and we use the RealPhoto60~\cite{yu2024supir} which contains 60 real-world degraded images without ground truth for evaluation. And we use DIV2K~\cite{timofte2017ntire} as the high-quality image database for retrieval and employ the image encoder of VGG16~\cite{simonyan2014very} as the feature extractor. As for the evaluation metrics, we use both the fidelity metrics containing PSNR and SSIM, as well as the perceptual metrics including LPIPS~\cite{zhang2018unreasonable}, NIQE~\cite{mittal2012making}, FID~\cite{heusel2017gans}, MUISQ~\cite{ke2021musiq}, and CLIPIQA~\cite{wang2023exploring}, to assess the performance of the different methods.

\begin{table}[!tb]
\centering
\caption{Quantitative comparison with state-of-the-art RefSR methods, GAN-based methods, and Diffusion-based methods on real-world image super-resolution. Our ReFIR achieves consistent performance improvements in both fidelity and perceptual quality.}
\label{tab:compare-real}
\setlength{\tabcolsep}{2.5pt}
\scalebox{0.88}{
\begin{tabular}{@{}l|ccccc|ccccc@{}}
\toprule
                            & \multicolumn{5}{c|}{\textbf{CUFED5}}        & \multicolumn{5}{c}{\textbf{WR-SR}}           \\
\multirow{-2}{*}{Method} &
  PSNR$\uparrow$ &
  SSIM$\uparrow$ &
  LPIPS$\downarrow$ &
  NIQE$\downarrow$ &
  FID$\downarrow$ &
  PSNR$\uparrow$ &
  SSIM$\uparrow$ &
  LPIPS$\downarrow$ &
  NIQE$\downarrow$ &
  FID$\downarrow$ \\ \midrule
C2-Matching~\cite{jiang2021robust}                 & 20.77 & 0.5169  & 0.7282 & 8.4438  & 282.43 & 22.63 & 0.5627  & 0.7177  & 8.3238  & 157.61 \\
DATSR~\cite{cao2022reference}                       & 20.75 & 0.5130  & 0.7301 & 8.6765  & 282.19 & 22.62 & 0.5620  & 0.7210  & 8.4329  & 157.54 \\
MrefSR~\cite{zhang2023lmr}                      & 20.84 & 0.5218  & 0.7853 & 9.6524  & 286.44 & 22.68 & 0.5703  & 0.7748  & 9.7742  & 156.57 \\ \midrule
BSRGAN~\cite{zhang2021designing}                      & 20.22 & 0.5256  & 0.4135 & 4.2204  & 203.17 & 22.07 & 0.5735  & 0.4073  & 3.8703  & 133.50 \\
Real-ESRGAN~\cite{wang2021real}                 & 20.31 & 0.5543  & 0.3698 & 3.8832  & 175.91 & 22.14 & 0.5974  & 0.3631  & 3.7001  & 97.88  \\
StableSR~\cite{wang2023stablesr}                    & 20.46 & 0.4480  & 0.6532 & 6.3433  & 292.69 & 21.22 & 0.4421  & 0.5899  & 5.2040  & 145.07 \\
DiffBIR~\cite{lin2023diffbir}                     & 19.76 & 0.4886  & 0.3820 & 3.5629  & 154.75 & 21.30 & 0.5284  & 0.3938  & 3.8736  & 76.05  \\
PASD~\cite{yang2023pasd}                        & 20.22 & 0.4959  & 0.5252 & 5.4828  & 208.64 & 21.12 & 0.5254  & 0.4292  & 4.2505  & 98.16  \\ \midrule
\rowcolor[HTML]{EFEFEF} 
SeeSR~\cite{wu2023seesr}  & 19.94 & 0.5195  & 0.3660 & 3.7912  & 142.92 & 21.73 & 0.5658  & 0.3501  & 4.0155  & 65.78  \\
\rowcolor[HTML]{EFEFEF} 
SeeSR+ReFIR                 & 20.32 & 0.5289  & 0.3338 & 3.7831  & 134.62 & 21.86 & 0.5664  & 0.3460  & 3.9089  & 61.22  \\
\rowcolor[HTML]{EFEFEF} 
\improvetit{$\Delta$improvement} & \improve{+0.38} & \improve{+0.0094} & \improve{+0.0322} & \improve{+0.0081} & \improve{+8.30}  & \improve{+0.13} & \improve{+0.0006} & \improve{+0.0041} & \improve{+0.1066} & \improve{+4.56}  \\ \midrule
\rowcolor[HTML]{EFEFEF} 
SUPIR                       & 18.97 & 0.4665  & 0.4807 & 4.5624  & 168.26 & 20.91 & 0.5426 & 0.3791  & 3.7587  & 75.85  \\
\rowcolor[HTML]{EFEFEF} 
SUPIR+ReFIR                 & 19.00 & 0.4729 & 0.4341 & 4.2085 & 148.69 & 21.02 & 0.5497  & 0.3785  & 3.7478  & 71.82  \\
\rowcolor[HTML]{EFEFEF} 
\improvetit{$\Delta$improvement} & \improve{+0.03}  & \improve{+0.0064}  & \improve{+0.0466} & \improve{+0.3539}  & \improve{+19.57}  & \improve{+0.11}  & \improve{+0.0071}  & \improve{+0.0006}  & \improve{+0.0109}  & \improve{+4.03}  \\ \bottomrule
\end{tabular}%
}
\end{table}

\begin{figure}[!t]
    \centering
    \includegraphics[width=1.\linewidth]{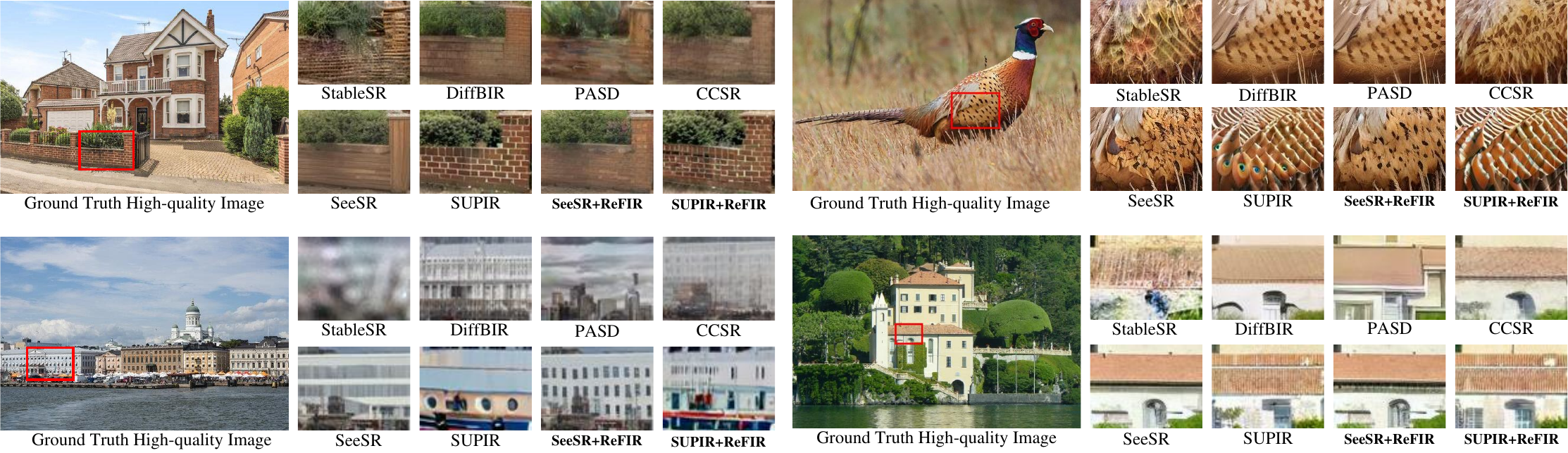}
    \caption{Quantitative comparison on RefSR dataset. The results using our ReFIR are \textbf{bolded}. Please zoom in for better visualization.}
    \label{fig:wrsr-viz}
    \vspace{-5mm}
\end{figure}

\noindent
\textbf{Implementation details.}
For a fair comparison, we use one reference image if not specified. Experiments with multiple reference images are given in \cref{sec:supp-multi-ref}. Following the common practice of existing LRMs~\cite{yu2024supir,yang2023pasd,wu2023seesr,wang2023stablesr}, the $I_{LQ}$ is up-sampled to the desired size using Bicubic before going through the LRMs. 
We use reflective padding to ensure the input size of $\mathcal{C}_T$ and $\mathcal{C}_S$ are the same. We use fixed random seeds for results reproducibility in all experiments. The hyperparameters of different baselines follow their original settings. We apply the proposed retrieval augmented restoration framework to two popular LRMs, namely SeeSR~\cite{wu2023seesr} and SUPIR~\cite{yu2024supir}, and denoted the models augmented with our ReFIR as ``SeeSR+ReFIR” and ``SUPIR+ReFIR”, respectively.

\subsection{Comparison to State-of-the-Arts}
\noindent \textbf{Restoration with ideal reference.}
We first compare on the RefSR dataset with real-world degradation. The compared methods includes state-of-the-art RefSR methods~\cite{jiang2021robust,cao2022reference,zhang2023lmr}, GAN-based methods~\cite{zhang2021designing,wang2021real}, and recent Diffusion-based methods~\cite{wang2023stablesr,yu2024supir,wu2023seesr,yang2023pasd,lin2023diffbir}. \cref{tab:compare-real} gives the results. It can be seen that our method brings significant gains in \textit{all} metrics on both fidelity (PSNR, SSIM) and perceptual quality (LPIPS, NIQE, FID) for the LRMs. Taking SUPIR as an example, our method brings a FID improvement of even 19.57 on the CUFED5 dataset. Moreover, similar performance gains can also be observed in SeeSR. For instance, equipping our ReFIR to SeeSR can lead to 0.38dB PSNR improvement, demonstrating the generalization of our ReFIR. It is noteworthy that the above superiority is obtained without any training or fine-tuning. Moreover, we also give visual comparisons in~\cref{fig:wrsr-viz}, and it can be seen that our method can generate details that are faithful to the original scene with the help of external knowledge from retrieved reference images. 

\begin{figure}[!t]
    \centering
    \includegraphics[width=1.\linewidth]{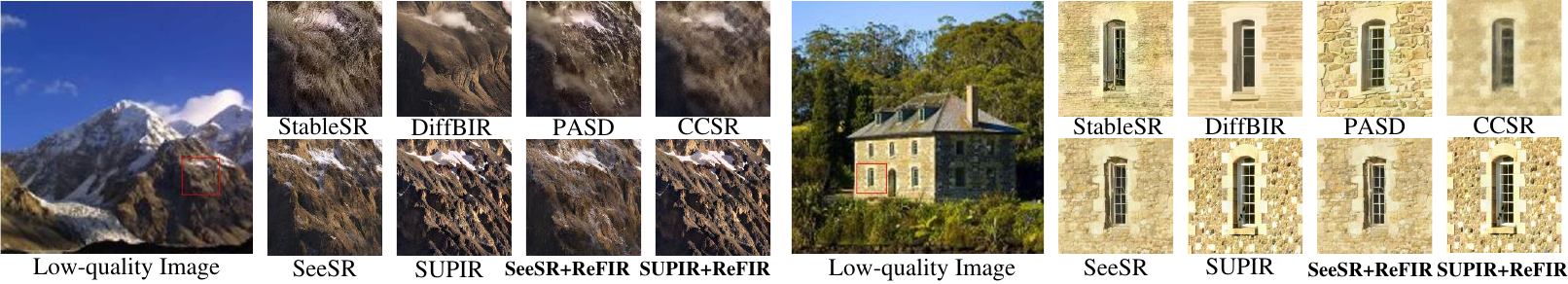}
    \vspace{-6mm}
    \caption{Quantitative comparison on RealPhoto dataset. More results are provided in \cref{sec:supp-additional-viz-res}.}
    \label{fig:realPhoto-main}
    \vspace{-6mm}
\end{figure}

\noindent \textbf{Restoration in the wild.}
The above experiments on RefSR datasets focus on utilizing the already provided reference images from the dataset, which applies when the user has relevant HQ images. In this section, we turn to more challenging scenarios in which the reference image has to be obtained by retrieval. Since the ground truth of RealPhoto datasets is unavailable, we use non-reference image quality assessment metrics, \textit{i.e.} NIQE, MUSIQ, and CLIPIQA for evaluation. As shown in \cref{tab:compare-realPhoto}, our approach continues to produce significant gains over its non-ReFIR counterparts. For instance, our SeeSR+ReFIR surpasses the original SeeSR by 0.2866 NIQE and 1.59 MUSIQ. Since the retrieved image can not serve as an ideal reference, the above favorable results demonstrate the robustness of our ReFIR in the face of real-world retrieved images. We also give quantitative results in~\cref{fig:realPhoto-main}. Even under severe real-world degradation, our method maintains good perceptual quality.

\noindent
\textbf{Complexity analysis.}
\cref{tab:supp-efficiency} gives the comparison of the computational complexity, including the number of parameters, GPU cost, and the inference latency. We also give the restoration performance for a more comprehensive comparison. As for the parameters, our ReFIR can facilitate both fidelity and realistic image restoration using the same \#param as the original base LRMs. For the GPU memory, since our ReFIR uses two images as input, \textit{i.e.}, one LQ image, and one reference image, the GPU cost will become larger than the original one. For instance, it rises 1.38 times the increase of SUPIR+ReFIR than the original SUPIR model. Moreover, the inference time also increases due to more inputs as well as the additional interaction between two chains. In the future, we will delve deep into the effective utilization of retrieved images while maintaining efficiency.

\begin{table}[!t]
\centering
\caption{Quantitative comparison on real-world degradation with RealPhoto datasets.}
\label{tab:compare-realPhoto}
\setlength{\tabcolsep}{2.5pt}
\scalebox{1.}{
\begin{tabular}{@{}l|cccccccc@{}}
\toprule
Metrics           & \makecell{StableSR \\ \cite{wang2023stablesr}} & \makecell{DiffBIR \\ \cite{lin2023diffbir}} & \makecell{PASD \\ \cite{yang2023pasd}}   & \makecell{CCSR \\ \cite{sun2023ccsr}}   & \makecell{SeeSR \\ \cite{wu2023seesr}}  & \makecell{SUPIR\\ \cite{yu2024supir}}  & \cellcolor{Gray}{\makecell{SeeSR+ReFIR \\ (Ours)}} & \cellcolor{Gray}{\makecell{SUPIR+ReFIR\\ (Ours)}} \\ \midrule
NIQE$\downarrow$  & 3.7695   & 2.8458  & 5.1603 & 5.5082 & 4.7432 & 3.5076 & \cellcolor{Gray}{4.4566\small{\improve{(+0.2866)}}}      & \cellcolor{Gray}{3.4593\small{\improve{(+0.0483)}}}      \\
MUSIQ$\uparrow$   & 51.95    & 65.20   & 49.01  & 32.26  & 55.54  & 59.84  & \cellcolor{Gray}{57.13\small{\improve{(+1.59)}}}       & \cellcolor{Gray}{60.49\small{\improve{(+0.65)}}}       \\
CLIPIQA$\uparrow$ & 0.6852   & 0.7845  & 0.5863 & 0.4568 & 0.6575 & 0.5692 & \cellcolor{Gray}{0.6732\small{\improve{(+0.0157)}}}      & \cellcolor{Gray}{0.5722\small{\improve{(+0.003)}}}      \\ \bottomrule
\end{tabular}%
}
\vspace{-3mm}
\end{table}

\subsection{Ablation Studies}

\begin{figure*}[!t]
\centering
 \includegraphics[width=\linewidth]{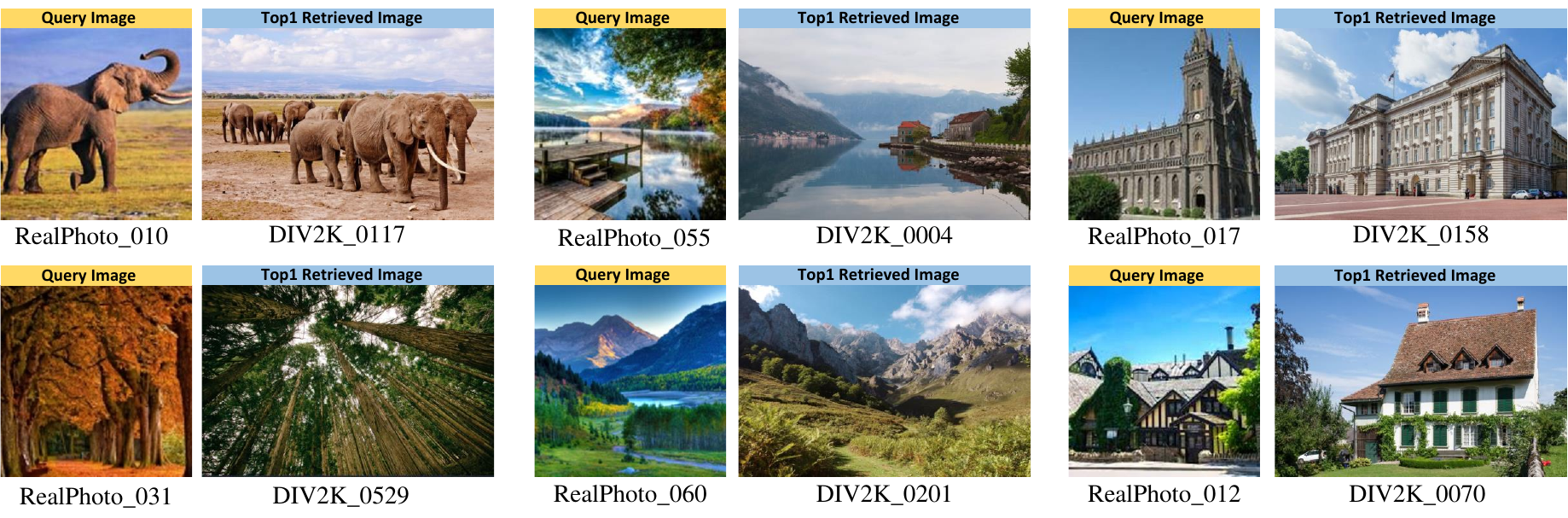}
 \vspace{-6mm}
 \caption{The retrieval results with RealPhoto60 dataset~\cite{yu2024supir} as the query images and DIV2K~\cite{timofte2017ntire} as the HQ image database.}
 \label{fig:supp-retrive-results}
\end{figure*}

\noindent
\textbf{Effectiveness of the reference retriever.}
\label{sec:supp-effctive-retriver}
In order to obtain content-relevant retrieved images, we present a simple but inference-efficient retriever $\mathcal{R}$ that uses the high-level semantic vectors from the pre-trained deep models for similarity matching in the high-quality image dataset $\mathcal{D}$. Despite the simple design, we here demonstrate its effectiveness in~\cref{fig:supp-retrive-results}. Since semantically consistent images usually contain similar textures, \textit{e.g.}, the texture in the first elephant image can help in the restoration of the LQ elephant image, and thus the proposed retriever can yield satisfactory retrieval results. Although texture-based retrieval may be a better choice for image restoration, it usually necessitates additional training of new retrieval models. For simplicity, we adopt semantic-based retrieval and leave the exploration of more advanced reference retrievers for future work.

\noindent
\textbf{Ablation on cross image injection.}
In the proposed cross image injection, we use separate attention (SA), spatial adaptive gating (SG), and distribution alignment (DA) for effective external knowledge injection. Here, we ablate to validate the effectiveness of different components. We use SUPIR+ReFIR as a representative on the CUFED5 dataset and use the scalar weighted sum when SG is removed. The results are shown in \cref{tab:ablation-components}. One can see that using fixed scalar weights instead of spatial adaptive gating results in a 0.18 NIQE drop. This is because not all pixels of the reference image are useful, and thus fine-grain gated mask is needed. Moreover, removing the distribution alignment also impairs performance, \textit{e.g.}, 4.36 FID drop, since the distribution of raw fusion results $O_{fuse}$ does not match $\mathcal{C}_T$, and directly inject $O_{fuse}$ to the denoising chain of $\mathcal{C}_T$ can cause sub-optimal results.

\begin{table}[!tb]
\centering
\caption{Comparison of model complexity before and after incorporating our ReFIR. We use an input image with the resolution of $2048 \times 2048$ to evaluate the GPU memory and the inference time on one single 80G NVIDIA A100 GPU.}
\label{tab:supp-efficiency}
\setlength{\tabcolsep}{2.5pt}
\scalebox{1.}{
\begin{tabular}{@{}l|cccc|ccc@{}}
\toprule
Method      & PSNR$\uparrow$  & SSIM$\uparrow$   & LPIPS$\downarrow$  & FID$\downarrow$    & \#param$\downarrow$ & GPU Memory$\downarrow$ & Inference Time$\downarrow$ \\ \midrule
SeeSR~\cite{wu2023seesr}       & 19.94 & 0.5195 & 0.3660 & 142.92 & 2.04B   & 24.4G      & 76.5s          \\
SUPIR~\cite{yu2024supir}       & 18.97 &  0.4665 & 0.4807 &  168.26 & 3.87B   & 37.3G      & 146.4s         \\
\rowcolor[HTML]{EFEFEF} 
SeeSR+ReFIR & 20.32 & 0.5289 &  0.3338 & 134.62  & 2.04B   & 40.9G      & 170.7s         \\
\rowcolor[HTML]{EFEFEF} 
SUPIR+RefIR & 19.00 & 0.4729  & 0.4341 & 148.69  & 3.87B   & 51.4G      & 322.8s         \\ \bottomrule
\end{tabular}%
}
\vspace{-3mm}
\end{table}

\begin{figure}[!tb]  
    \vspace{-3mm}
    \begin{minipage}[t]{0.48\linewidth}
        \centering
        \captionsetup{width=0.96\linewidth}
        \captionof{table}{Effectiveness of different components in cross image injection.}
        \label{tab:ablation-components}
        \vspace{0mm}
        \setlength{\tabcolsep}{1.8pt}
        \scalebox{0.799}{
        \begin{tabular}{@{}ccc|cccc@{}}
\toprule
SA         & SG         & DA         & PSNR$\uparrow$  & SSIM$\uparrow$   & NIQE$\downarrow$   & FID$\downarrow$    \\ \midrule
           &            &            & 18.97 & 0.4665 & 4.5624 & 168.26 \\
\checkmark &            & \checkmark & 19.09 & 0.4799 & 4.3893 & 150.81 \\
\checkmark & \checkmark &            & 19.12 & 0.4724 & 4.2275 & 153.05 \\
\rowcolor[HTML]{EFEFEF} 
\checkmark & \checkmark & \checkmark & 19.00 & 0.4729 & 4.2085 & 148.69 \\ \bottomrule
\end{tabular}%
        }
    \end{minipage}
    \begin{minipage}[t]{0.48\linewidth}
    \centering
    \captionsetup{width=0.9\linewidth}
        \captionof{table}{Ablation experiments on different positions of cross-image injection.}
        \label{tab:ablation-inject-position}
        \vspace{0mm}
        \setlength{\tabcolsep}{1.8pt}
        \scalebox{0.799}{
\begin{tabular}{@{}l|cccc@{}}
\toprule
Injection Position & PSNR$\uparrow$  & SSIM$\uparrow$   & NIQE$\downarrow$   & FID$\downarrow$   \\ \midrule
No Injection       & 18.97 & 0.4665 & 4.5624 & 168.26 \\
Encoder\&Decoder   & 19.09 & 0.4773 & 4.5241 & 153.43 \\
Encoder Only       & 19.08 & 0.4689 & 4.6977 & 168.26 \\
\rowcolor[HTML]{EFEFEF} 
Decoder Only       & 19.00 & 0.4729 & 4.2085 & 148.69 \\ \bottomrule
\end{tabular}%
        }
    \end{minipage}
    \vspace{-3mm}
\end{figure}

\begin{wrapfigure}{r}{0.35\linewidth}
  \vspace{-1mm}
 
 \centering
  \includegraphics[width=0.8\linewidth]{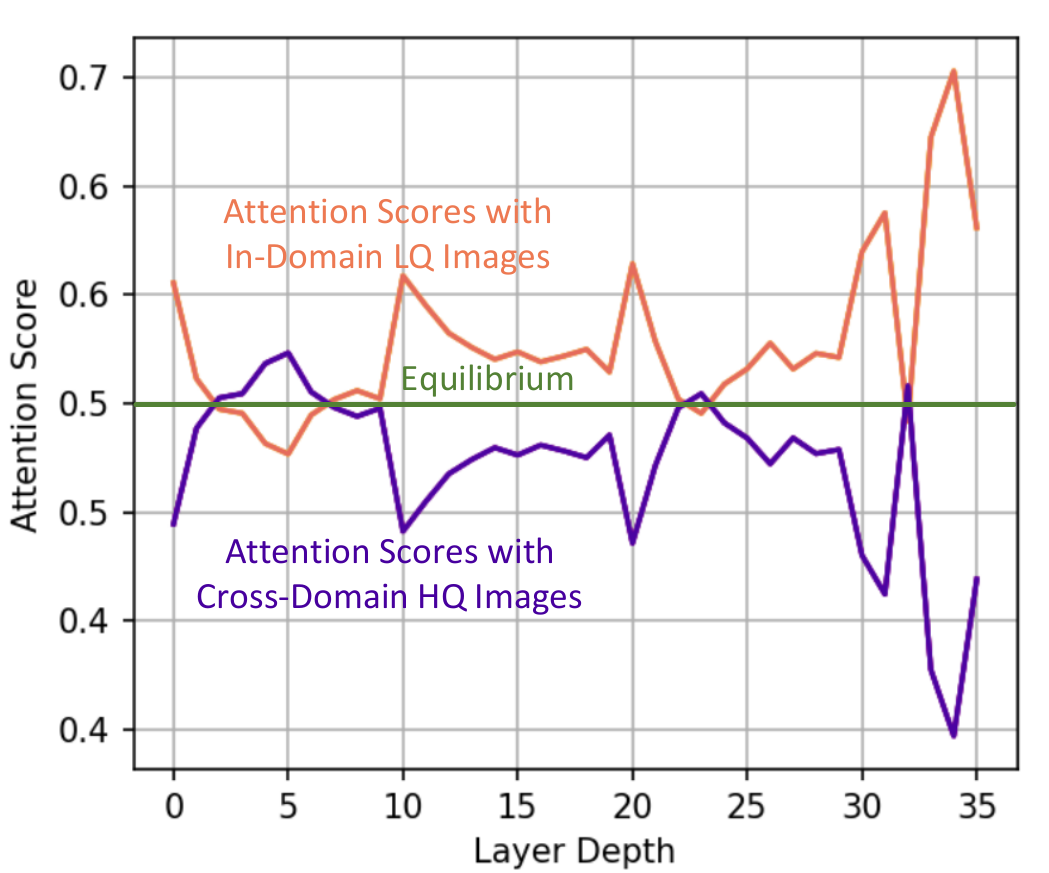}
  \vspace{-3mm}
  \caption{\small{The normalized attention scores are obtained by averaging all samples and all time steps.}}
   \label{fig:domain-prefernce}
\vspace{-4mm}
\end{wrapfigure}

\paragraph{Domain preference problem.}
\label{sec:diss-separate-attention}
The motivation behind the proposed separate attention is to address the domain preference problem, \textit{i.e.}, the attention in $\mathcal{C}_T$ will prefer to use latent from the same chain even though the latent from $\mathcal{C}_S$ is more helpful for reconstruction. To verify the existence of the domain preference, we use the ground truth $I_{HR}$ as the input of $\mathcal{C}_S$ and compute the normalized attention scores between $Q_T$ and $K_T$, $Q_T$ and $K_S$. It can be seen in \cref{fig:domain-prefernce} that even using the spatially strictly aligned $I_{HQ}$ as the reference, $Q_T$ still has significantly high attention for the latent from the same chain, indicating that the domain preference problem interferes with the $\mathcal{C}_T$'s utilization of external knowledge in $\mathcal{C}_S$. By contrast, the proposed separate attention can effectively mitigate this problem by forcing the $Q_T$ to separately attend $K_T$ and $K_S$.

\noindent
\textbf{Other choices on injection position.}
In \cref{sec:observation}, we find the diffusion decoder is responsible for restoring textures. Based on this observation we propose to apply cross-image injection on the UNet decoder. Here, we ablate to analyze the impact of different cross-image injection positions. The results are shown in~\cref{tab:ablation-inject-position}. It can be seen that performing cross-image injection only on the encoder will cause 19.57 FID drops. This is because the encoder focuses on the structure reconstruction, thus transferring the structure of $\mathcal{C}_S$ will destroy the layout of the $\mathcal{C}_T$. Moreover, performing injection only in the decoder achieves the best results since it can transfer the high-quality textures from the $\mathcal{C}_S$. Due to the page limit, more ablation experiments can be seen in~\cref{sec:supp-ablation}.

\subsection{Discussions}

\begin{figure}[!t]
    \centering
    \includegraphics[width=1.\linewidth]{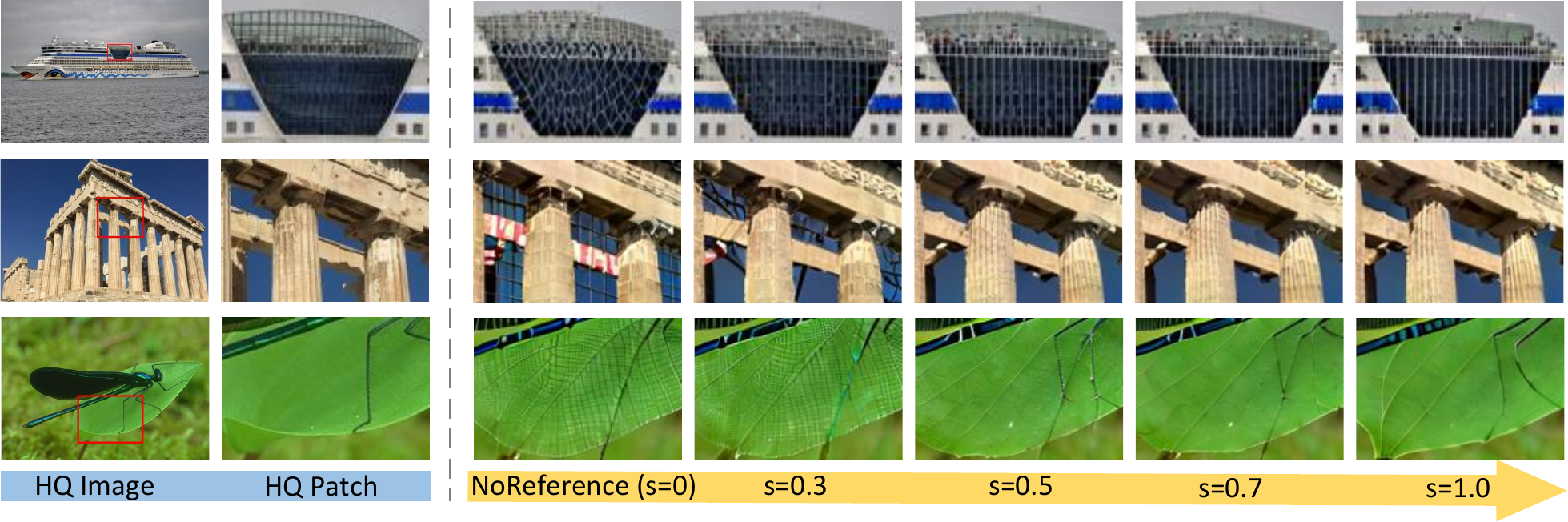}
    \vspace{-7mm}
    \caption{Ablation visualization on the control scale $s$. As $s$ increases, the LRM utilizes the external knowledge from retrieved reference images to mitigate hallucination. Zoom in for better effects.}
    \label{fig:abaltion-scale}
    \vspace{-2mm}
\end{figure}

\noindent
\textbf{What is the impact of the control scale?}
The scale $s$ in \cref{eq:scale-s} can control the extent to which the LRMs use external knowledge from the retrieved reference image for restoration. Here, we conduct an ablation study to explore the effect of $s$. The results are shown in~\cref{fig:abaltion-scale}. It can be seen that when $s$ takes smaller values, the model mainly uses the internal knowledge embedded in its own parameters, which can make the model hallucinate when the degradation is severe. For example, the model produces incorrect textures when $s=0$. As $s$ increases, the model starts to use external knowledge from the retrieved reference image, from which the model's hallucination problem can be alleviated. We also provide quantitative ablation experiments on $s$ in~\cref{sec:supp-ablation-quant-scale}.

\begin{figure}[!tb]  
    \begin{minipage}[t]{0.48\linewidth}
        \centering
        \captionsetup{width=0.96\linewidth}
        \captionof{table}{The performance impact of reference images. NoRef means no reference image is used. HQRef denotes the corresponding $I_{HQ}$ is used as the reference. SelfRef represents using $\times 4 $ bicubic upsampling of $I_{LQ}$ for reference. Random means randomly selecting a high-quality image as the reference.}
        \label{tab:up-bound}
        \vspace{1mm}
        \setlength{\tabcolsep}{1.8pt}
        \scalebox{0.799}{
        \begin{tabular}{@{}l|ccccc@{}}
        \toprule
        Settings & PSNR$\uparrow$  & SSIM$\uparrow$   & LPIPS$\downarrow$  & NIQE$\downarrow$   & FID$\downarrow$    \\ \midrule
        NoRef    & 18.97 & 0.4665 & 0.4807 & 4.5624 & 168.26 \\
        HQRef    & 19,41 & 0.5033 & 0.3928 & 4.0764 & 137.52 \\
        SelfRef  & 19.16 & 0.4795 & 0.4761 & 4.5501 & 163.94 \\
        Random   & 19.53 & 0.5138 & 0.5354 & 5.3796 & 223.47 \\
        \rowcolor{Gray} 
        Baseline & 19.00 & 0.4729 & 0.4341 & 4.2085 & 148.69 \\ \bottomrule
        \end{tabular}%
        }
    \end{minipage}
    \begin{minipage}[t]{0.48\linewidth}
    \centering
    \captionsetup{width=0.9\linewidth}
    \caption{An explanation of how the proposed retrieval augmented framework affects the restoration process of existing LRMs.}
      \label{fig:supp-diff-manifold}
    \vspace{-2mm}
    \includegraphics[width=0.8\columnwidth]{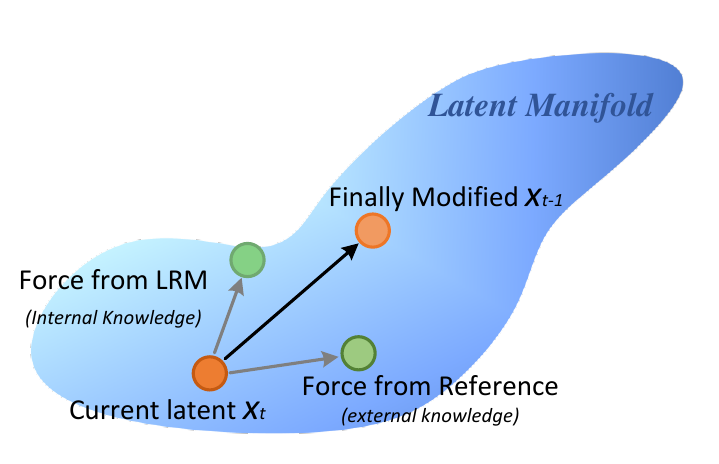}
    \end{minipage}
    \vspace{-4mm}
\end{figure}

\noindent
\textbf{How much do the reference images affect performance?}
In the proposed framework, the retrieved images $\mathbf{I_{R}}$ is crucial in alleviating hallucinations. Here, we try to answer the role of $\mathbf{I_{R}}$ during restoration process, by manually controlling different types of retrieved images.
As shown in \cref{tab:up-bound}, we find that using the exact ground truth $I_{HQ}$ as the $\mathbf{I_R}$ can further improve the performance, which can be seen as an ideal up-bound. Interestingly, using $I_{LQ}$ itself as its own retrieved image instead brings a slight improvement compared with no retrieval, which we attribute to the regularization effect from the distribution alignment strategy. Finally, randomly selecting a high-quality reference image even resulted in a huge performance degradation, suggesting that the content correlation is more important than the image quality for a favorable retrieved reference image.

\noindent \textbf{How does the proposed ReFIR work?}
Extensive experiments have shown the state-of-the-art performance of our ReFIR. However, it seems not straightforward to understand how the retrieved reference images influence the image restoration process of the original LRMs. Here, we give an intuitive explanation. As shown in \cref{fig:supp-diff-manifold}, for the latent at the $t$-th time step on the latent manifold, there are two forces in different directions pulling it to produce the latent at the next $t-1$-th time step. One force is from the internal knowledge of frozen weights in LRMs, and the other is the external knowledge from the retrieved reference image through the proposed cross image injection mechanism. These two forces ultimately determine the latent of the next time step. Therefore, a restored image from our ReFIR can utilize both the internal knowledge in the original LRMs as well as the external knowledge in the retrieved image, thus alleviating the hallucination of the LRMs.

\section{Conclusion}
This paper presents ReFIR, a training-free and generic framework that can alleviate the hallucination of LRMs to facilitate high-fidelity and photo-realistic restoration results through retrieval augmentation. We introduce the nearest neighbor lookup as a simple retriever to obtain relevant high-quality images and further propose the cross-image injection which employs separate attention to transfer knowledge while avoiding the domain preference problem, the spatial adaptive gating to address the spatial misalignment, and the distribution alignment to mitigate the domain gap during injection. Through expanding the knowledge boundary using the additional external knowledge from retrieved images, our ReFIR exhibits significant improvements on both fidelity and perceptual quality, as demonstrated through extensive qualitative and quantitative evaluations. Moreover, with its training-free and generic nature, our ReFIR can be easily applied to multiple LRMs.

\clearpage

\section*{Acknowledgements}
\vspace{-3mm}
This work is supported in part by the National Natural Science Foundation of China, under Grant (62302309,62171248), Shenzhen Science and Technology Program (JCYJ20220818101014030, JCYJ20220818101012025), and the PCNL KEY project (PCL2023AS6-1).

{
  \small
  \bibliographystyle{unsrt}
  \bibliography{neurips_2024}
}
\clearpage
\appendix
\section*{\Large{Appendix}}
\label{sec:Appendix}





\section{Adaptive Multi-reference Injection}
\label{sec:supp-multi-ref}

\paragraph{Technical details.}
In the main paper, we mainly focus on the case of one single retrieved image. However, in practice, there may be multiple available reference images at hand, and using multiple reference images for resemblance could intuitively gain better performance. To this end, we extend the original cross-image injection to allow to incorporation of multiple reference images for reconstruction. 
Our key idea is to modify the scale factor $s$ in \cref{eq:scale-s} from a scalar into a vector: $\mathbf{s}=\{s_1, s_2, \cdots, s_k\}$, where $\sum s_n=1$. Each $s_n$ can be obtained by computing the cosine similarity between $I_{LQ}$ and the corresponding $n$-th retrieved image in $\mathbf{I_R}$ followed by Softmax normalization. Then we can modify the original single-reference
cross-image injection of \cref{eq:scale-s} to the following multi-reference version:
\begin{equation}
O_{fuse} = (\mathbf{1}-\sum_{n=1}^{k}s_n \mathcal{M}_n)\otimes O_{inter} + (\sum_{n=1}^{k}s_n \mathcal{M}_n)\otimes O_{intra},
\end{equation}
\noindent
where $\mathcal{M}_n$ denotes the gated mask of the $n$-th reference image.

\paragraph{Experiments with multiple reference images.}
For experiments with multiple reference images, we use SUPIR+ReFIR as a representative. Since the CUFED5 dataset contains multiple reference images, we directly use the provided images as the retrieved reference for reproducibility. \cref{tab:ablation-multi-ref} gives the results. It can be seen that using multiple reference images produces better results than one single reference image, \textit{e.g.} the 2.08 improvement in FID. However, it is worth noting that the marginal gain from adding reference images is diminishing, accompanied by a notable increase in computational cost. Therefore, in practice, we use one single reference image to balance the model performance and inference efficiency.

\section{More Discussions}
\label{sec:supp-discuss}

\paragraph{Difference from the other methods.}
\label{sec:supp-difference-RefSR}
Our ReFIR uses retrieved images as the reference for high-fidelity restoration. Despite both RefSR methods and ours appears the reference image, we would like to clarify the difference between our ReFIR and previous RefSR methods. \textbf{Firstly}, current RefSR models~\cite{jiang2021robust,cao2022reference,zhang2023lmr} are typically small-scale (\#param <50M) and use simple Bicubic degradation, while our ReFIR focuses on the recent diffusion-based large-scale restoration model (\#param >1B) for more challenging real-world SR. \textbf{Secondly}, most RefSR methods can only use one reference image and even fail to work in the absence of reference images, by contrast, our ReFIR can flexibly use $0\sim k$ images. \textbf{Thirdly}, different from RefSR models that require training, our method can be applied in various LRMs in a training-free manner.

\begin{wrapfigure}{r}{0.40\linewidth}
    \centering
    \captionof{table}{\small{Results of all LR images using the fallback strategies.}}
    \label{tab:suppl-perfromance-extreme}
    \setlength{\tabcolsep}{2.5pt}
    \scalebox{0.9}{
    \begin{tabular}{@{}l|ccc@{}}
    \toprule
    setup       & NIQE$\downarrow$   & MUSIQ$\uparrow$                           & CLIPIQA$\uparrow$ \\ \midrule
    origin\_lrm  & 4.7432 & 55.54                                                & 0.6575  \\
    gen\_ref     & 4.6923 & 55.98                                                & 0.6602  \\
    ada\_gen\_ref & 4.3464 & 57.68 & 0.6942  \\
    \rowcolor[HTML]{EFEFEF} 
    ReFIR       & 4.4986 & 57.01                                                & 0.6759  \\ \bottomrule
    \end{tabular}%
    }
    \vspace{-3mm}
\end{wrapfigure}

\paragraph{Performance under extreme conditions.}
Since our ReFIR relies on the retrieved images, it is interesting to explore extreme situations when highly relevant and high-quality reference images are scarce or even unavailable. To this end, we introduce the fallback strategies to handle this situation. Specifically, since our method does not modify the parameters of LRMs, we can directly use the original inference pipeline of the LRM without using reference images. We denote this as $\mathrm{origin\_lrm}$. In addition, we also use the BLIP model to caption the LR image to obtain the text prompt, which will then be fed into the StableDiffusion2.0 model to generate semantic-similar high-quality images as the reference. We denote this as $\mathrm{gen\_ref}$. We use SeeSR~\cite{wu2023seesr} as a representative, on the real-world degradation dataset RealPhoto60~\cite{yu2024supir}. We first give the results in which all LR images adopt the fallback strategies in \cref{tab:suppl-perfromance-extreme}. It can be seen that using the SD2.0 generated images as the fallback image can bring slightly improvement compared with noReference. After that, we further develop task-oriented adaptive strategies to enhance the performance of ReFIR in real-world scenarios. In detial, we respectively use the retrieved images and the $\mathrm{gen\_ref}$ to generate the results. And then we select the one with a larger task score as the final result. We denote it as $\mathrm{ada\_gen\_ref}$. From \cref{tab:suppl-perfromance-extreme}, it can be seen that the task-oriented strategy achieves a significant performance improvement against previous ReFIR baselines, \textit{e.g.}, 0.0183 CLIPIQA improvements, due to the fact that it works in the output end. However, this setup is accompanied by a larger inference time, and further acceleration on this fallback strategies can be an promising future work.

\noindent
\textbf{Computational overhead from retrieval and attention modification.}
Since we employ additional Ref images as input and modify the attention layers, we adiscuss the impact of these trchnuques on the inferenve efficiency. First, in order to reduce the computational overhead of the retrieval process, we pre-calculated the feature vectors of all images in the retrieval database before inference. Furthermore, the cosine similarity between the LR image vectors and all retrieval vectors is computed in parallel. These strategy results in an almost negligible (less than 3\% inference time) cost of computational overhead. Second, the modification of self-attention layers only happens in the last 20 timestep in the decoder layers, \textit{i.e.}, only 12\% attention layers are modified while the left is kept intact. These analysis is also supported by practice, in which we find these two process only take up <5\% inference time, with most computational cost coming from the original LRM. Future LRM acceleration (\textit{e.g.} pruning, quantization, one-step diffusion) will benefit our ReFIR, and we will explore more efficient implementation in the future.

\begin{table}[!t]
\centering
\caption{Experiments on extending to use multiple retrieved images for restoration. The inference time is evaluated on A100 GPU.}
\label{tab:ablation-multi-ref}
\resizebox{0.9\columnwidth}{!}{%
\begin{tabular}{@{}l|cc|ccccc@{}}
\toprule
settings & GPU Memory$\downarrow$ & Inference Time$\downarrow$ & PSNR$\uparrow$  & SSIM$\uparrow$   & LPIPS$\downarrow$  & NIQE$\downarrow$   & FID$\downarrow$    \\ \midrule
NoRef    & 37.3G      & 146.4s         & 18.97 & 0.4665 & 0.4807 & 4.5624 & 168.26 \\
OneRef   & 51.4G      & 322.8s         & 18.86 & 0.4623 & 0.4492 & 4.2317 & 156.10 \\
TwoRef   & 65.5G       & 499.2s         & 18.78 & 0.4676 & 0.4296 & 4.2315 & 154.02 \\ \bottomrule
\end{tabular}%
}
\vspace{-3mm}
\end{table}

\begin{figure*}[!t]
\centering
 \includegraphics[width=0.9\linewidth]{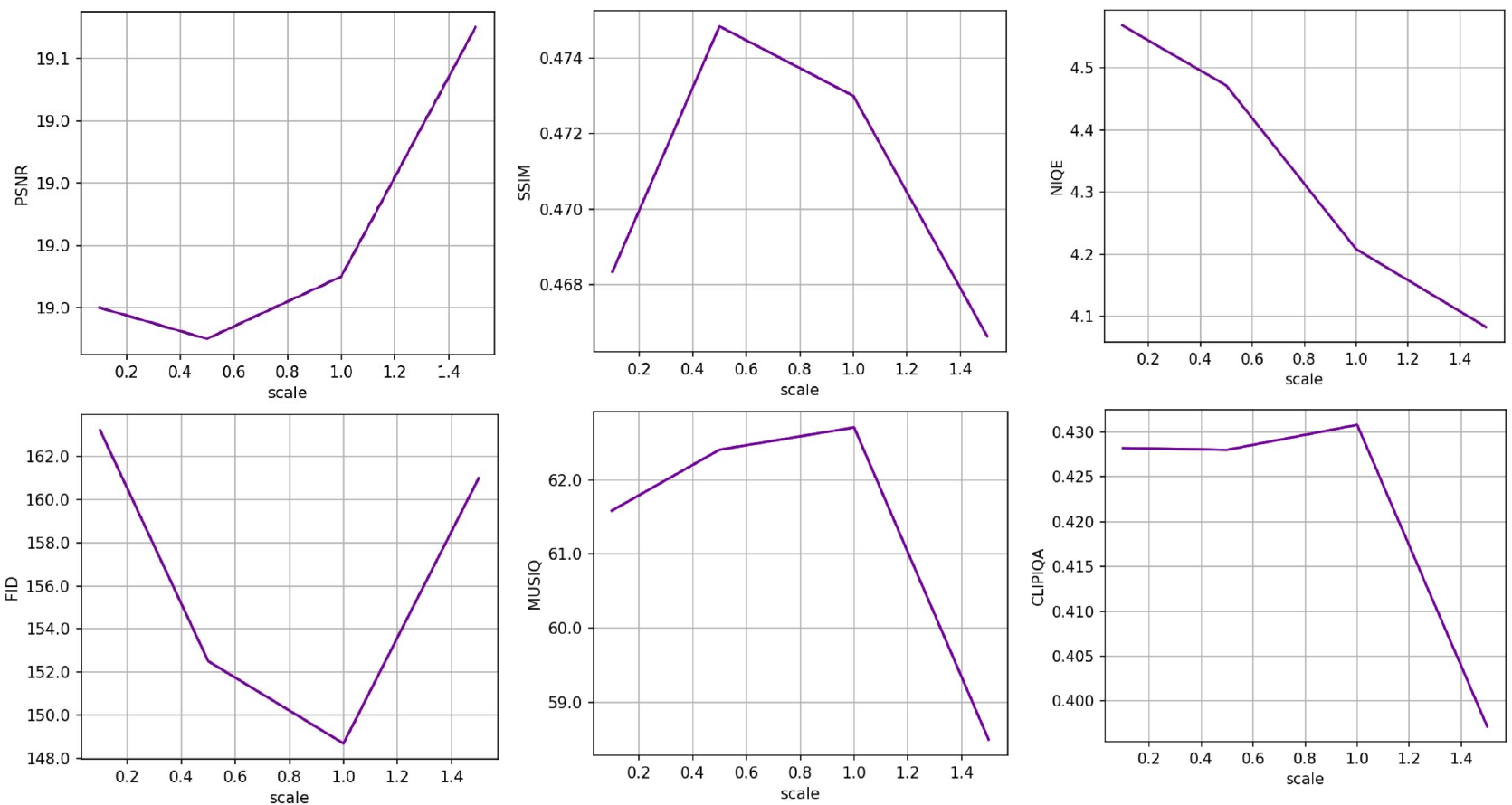}
 \caption{Quantitative ablation results on the control scales using SUPIR+ReFIR on CUFED5.}
 \label{fig:supp-ablation-scale}
 \vspace{-1mm}
\end{figure*}

\paragraph{Why use the self-attention as the external knowledge?}
In the proposed cross-image injection, we use the features of the self-attention layer of the $\mathcal{C}_S$'s decoder as external knowledge to guide $\mathcal{C}_T$ to produce textures faithful to the original scene. Here, we give the reason behind this. \textbf{Firstly}, previous image-to-image efforts~\cite{cao2023masactrl,gu2024photoswap,zhang2024real,hertz2022prompt}, \textit{e.g.}, image editing, has demonstrated through extensive experiments that the self-attention layer of the diffusion model contains important spatial correlations in images, which inspired us to follow this clue to utilize this prior. \textbf{Secondly}, leveraging the attention mechanism allows $\mathcal{C}_T$ to query features in $\mathcal{C}_S$ without any training, whereas using features from other parts of $\mathcal{C}_S$ may require introducing additional training.

\paragraph{What about the quality of cross-image attention?}
In the proposed cross-image injection, the inter-chain attention is used to perform attention between $Q_T$ and $K_S$. Considering the domain gap between $\mathcal{C}_T$ and $\mathcal{C}_{S}$ due to the input quality difference, one may ask whether the results of the inter-chain attention are meaningful. Here, We visualize the attention map to validate the effectiveness of the inter-chain attention (see~\cref{fig:supp-cross-attn}). It can be seen that for a given query pixel query in $\mathcal{C}_T$, the inter-chain attention can effectively overcome the spatial misalignment, and find relevant pixel features in $\mathcal{C}_S$ for reference.

\section{More Ablation Results}
\label{sec:supp-ablation}

\paragraph{Quantitative ablation on the control scale.}
\label{sec:supp-ablation-quant-scale}
We also provide quantitative ablation results on the control scale $s$ in~\cref{fig:supp-ablation-scale}. It can be seen that when $s$ is too small, the LRM will mainly use the knowledge contained within its parameters to restore high-quality images, which can lead to performance degradation due to the hallucination problem. On the other hand, when $s$ is too large, the LRM will overuse the content in the retrieved reference image, thus producing patterns that are not present in the original LQ image. In practice, we adopt a moderate $s=0.5$ to trade off the hallucination and the overuse of the reference image.

\begin{wrapfigure}{r}{0.40\linewidth}
    \centering
    \captionof{table}{\small{Ablation experiments on different cross image injection designs.}}
     \label{tab:supp-ablation-inject-manner}
    \setlength{\tabcolsep}{2.5pt}
    \scalebox{0.9}{
    \begin{tabular}{@{}l|cccc@{}}
        \toprule
        settings & PSNR$\uparrow$ & SSIM$\uparrow$   & NIQE$\downarrow$ & FID$\downarrow$    \\ \midrule
        replace  & 18.84 & 0.4385 & 4.26 & 182.82 \\
        concat   & 18.89 & 0.4691 & 4.19 & 156.03 \\
        \rowcolor[HTML]{EFEFEF} 
        baseline & 19.00 & 0.4729 & 4.21 & 148.69 \\ \bottomrule
        \end{tabular}%
    }
\end{wrapfigure} 

\paragraph{Other choices for cross image injection.}
The proposed cross image injection mitigates the domain preference problem by using separate attention to promote latent in $\mathcal{C}_T$ to attend $\mathcal{C}_S$. Here, we conduct ablation to study the impact of different design choices of cross image injection. As shown in~\cref{tab:supp-ablation-inject-manner}, directly replacing the original self-attention results from $O_{intra}$ in $\mathcal{C}_T$ with corresponding latent in $\mathcal{C}_S$ causes severe performance degradation, due to the significant loss of original knowledge in $\mathcal{C}_T$.
In addition, using $Q_T$ to query the concatenation results of $K_T$ and $K_S$ also causes a performance drop, which further confirms that the domain preference problem, \textit{i.e.}, $Q_T$ prefers to use latent from the same chain $\mathcal{C}_T$, even though $\mathcal{C}_S$ is more helpful for reconstruction.

\section{Statistical Significance on Performance}
\label{sec:supp-static-significant}
In~\cref{tab:compare-real} of the main paper, we give the performance gains of incorporating the proposed ReFIR into the existing LRMs. Considering the randomness of the generative models, we give the performance fluctuations of ReFIR under multiple trials with exactly the same experimental setting and random seed. The results are given in~\cref{tab:performace-fluctuation}. It can be seen that the randomness of the diffusion-based generative model is very small when using a fixed seed, reducing the disturbance from noise errors for evaluation. In addition, we further use hypothesis testing to verify the significance of performance gains, and the test results reject the original hypothesis H0 at 95\% confidence level on all metrics and datasets, indicating that the performance gains from the proposed method are statistically significant.

\section{Extension to Specific Restoration Scenarios}
An important application of our method is in scenarios with high fidelity demands, such as scene text images with a specific stylistic structure, or face images with identity preservation, and here we preliminarily explore the application of the proposed ReFIR to real-world face image restoration. The results are given in~\cref{fig:supp-faceRestore}. It can be seen that by using a high quality image of a specific person's identity as a reference, the resulting restoration results can better preserve the person's attributes. However, it should be noted that this experiment is just a preliminary attempt, and we will leave the further improvement of our ReFIR for specific downstream restoration tasks for future work.

\section{Limitation and Future Works}
\label{sec:supp-limitation}
Although the proposed ReFIR can effectively mitigate the hallucination of LRMs by introducing external knowledge from retrieved reference images, the proposed framework can be further improved in the following aspects. First, since the computational complexity of the current LRMs is costly, the computational complexity will be further increased when using the proposed method, which may hinder the use of resource-constrained mobile devices. With the advent of accelerated diffusion-based image restoration methods in the future, we believe that the proposed method can further improve its efficiency. In addition, this paper proposes a simple retriever based on semantic vector matching for presentation. With the development of image retrieval techniques~\cite{yan2020deep,huang2023vop}, designing specialized retrievers, \textit{e.g}., using textures as key matching cues, will further improve the performance. Finally, for some slightly degraded images, which can already be handled well by only using the internal knowledge of the LRMs, designing hyper-networks to adaptively decide whether to use retrieval augmentation or not is also promising. We leave the above considerations for future work.

\begin{table}[!t]
\centering
\caption{Performance fluctuations under different experiment trials. We use ten trails to obtain a stable fluctuation range.}
\label{tab:performace-fluctuation}
\resizebox{\columnwidth}{!}{%
\begin{tabular}{@{}l|ccccc|ccccc@{}}
\toprule
\multirow{2}{*}{settings} & \multicolumn{5}{c|}{\textbf{CUDED5}}                          & \multicolumn{5}{c}{\textbf{WR-SR}}                        \\
                          & PSNR$\uparrow$       & SSIM$\uparrow$        & LPIPS$\downarrow$      & NIQE$\downarrow$       & FID$\downarrow$      & PSNR$\uparrow$      & SSIM$\uparrow$       & LPIPS$\downarrow$       & NIQE$\downarrow$     & FID$\downarrow$     \\ \midrule
SeeSR+ReFIR               & 20.32      & 0.5289      & 0.3338     & 3.7831     & 134.62   & 21.86     &  0.5664     & 0.3460      & 3.9089  & 61.22   \\
Numerical fluctuations    & $\pm$0.0013 & $\pm$0.0001 & $\pm$0.0002 & $\pm$0.0001 & $\pm$0.03 & $\pm$0.001 & $\pm$ 0.0001 & $\pm$0.0002 & $\pm$0.0001 & $\pm$ 0.02 \\ \midrule
SUPIR+ReFIR               & 19.00     & 0.4729      & 0.4341     & 4.2085     & 148.69   & 21.02     &  0.5497     &  0.3785     & 3.7478    & 71.82   \\
Numerical fluctuations    & $\pm$0.005 & $\pm$0.0003 & $\pm$0.001 & $\pm$0.012 & $\pm$0.5 & $\pm$0.002 & $\pm$0.0003 & $\pm$0.0006 & $\pm$0.007 & $\pm$0.3 \\ \bottomrule
\end{tabular}%
    }
\end{table}

\section{Broader Impact}
\label{sec:supp-broader-impact}
The development of our ReFIR offers significant positive societal impacts, including advancements in medical imaging, historical preservation, and media restoration by enhancing the fidelity and realism of image restoration. However, it also poses potential negative societal impacts, such as the misuse of improved restoration capabilities for generating disinformation, deepfakes, and surveillance, raising ethical concerns about privacy, security, and fairness. To mitigate these risks, implementing safeguards like gated releases of models, monitoring mechanisms, and transparency in model training and deployment is crucial. Continuous ethical evaluation and adherence to strict guidelines are essential to prevent potential harms.

\section{Additional Visual Results}
\label{sec:supp-additional-viz-res}

\begin{figure*}[p]
\centering
 \includegraphics[width=1\linewidth]{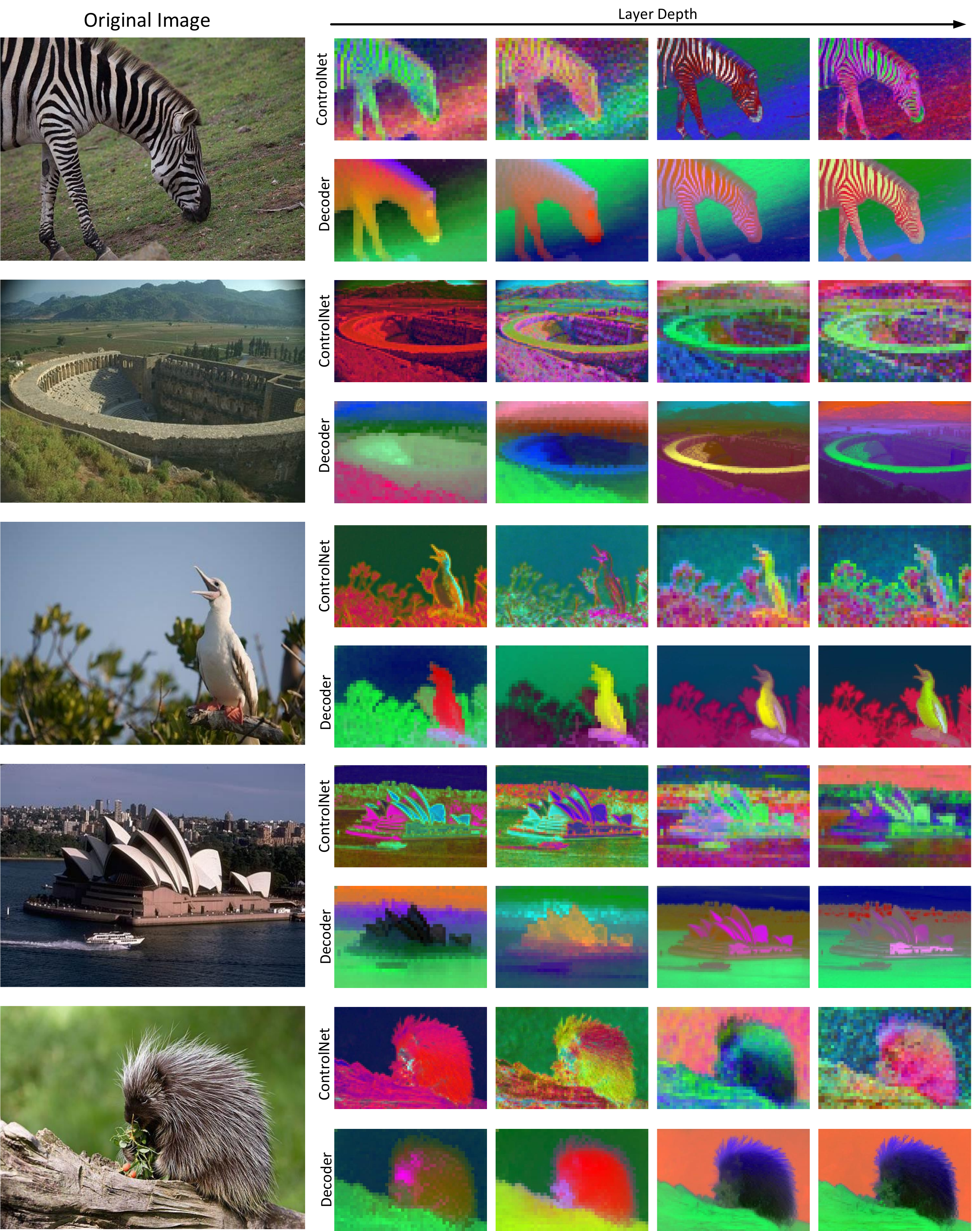}
 \caption{Additional visualization on the top three principal components of the self-attention layer latent of PCA. The latent is extracted from the first self-attention layer within blocks of the control net and unet decoder.}
 \label{fig:supp-pca}
\end{figure*}

\begin{figure*}[p]
    \centering
     \includegraphics[width=1\linewidth]{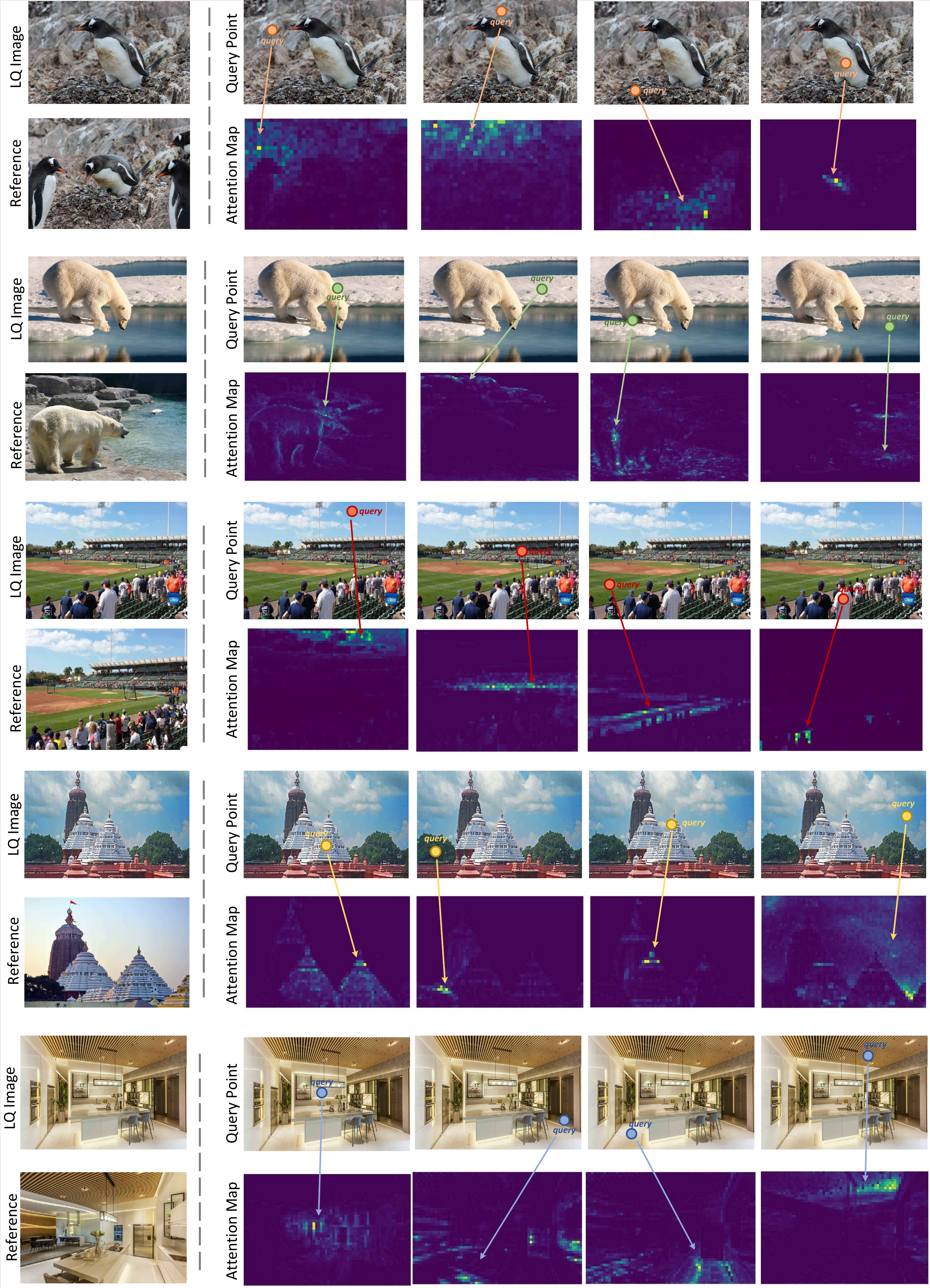}
     \caption{Additional visualization on the attention maps from the cross image injection. It can be seen that the query pixel in the $\mathcal{C}_T$ can well attend similarly region from $\mathcal{C}_S$.}
     \label{fig:supp-cross-attn}
     \vspace{-2mm}
\end{figure*}

\begin{figure*}[p]
\centering
 \includegraphics[width=1\linewidth]{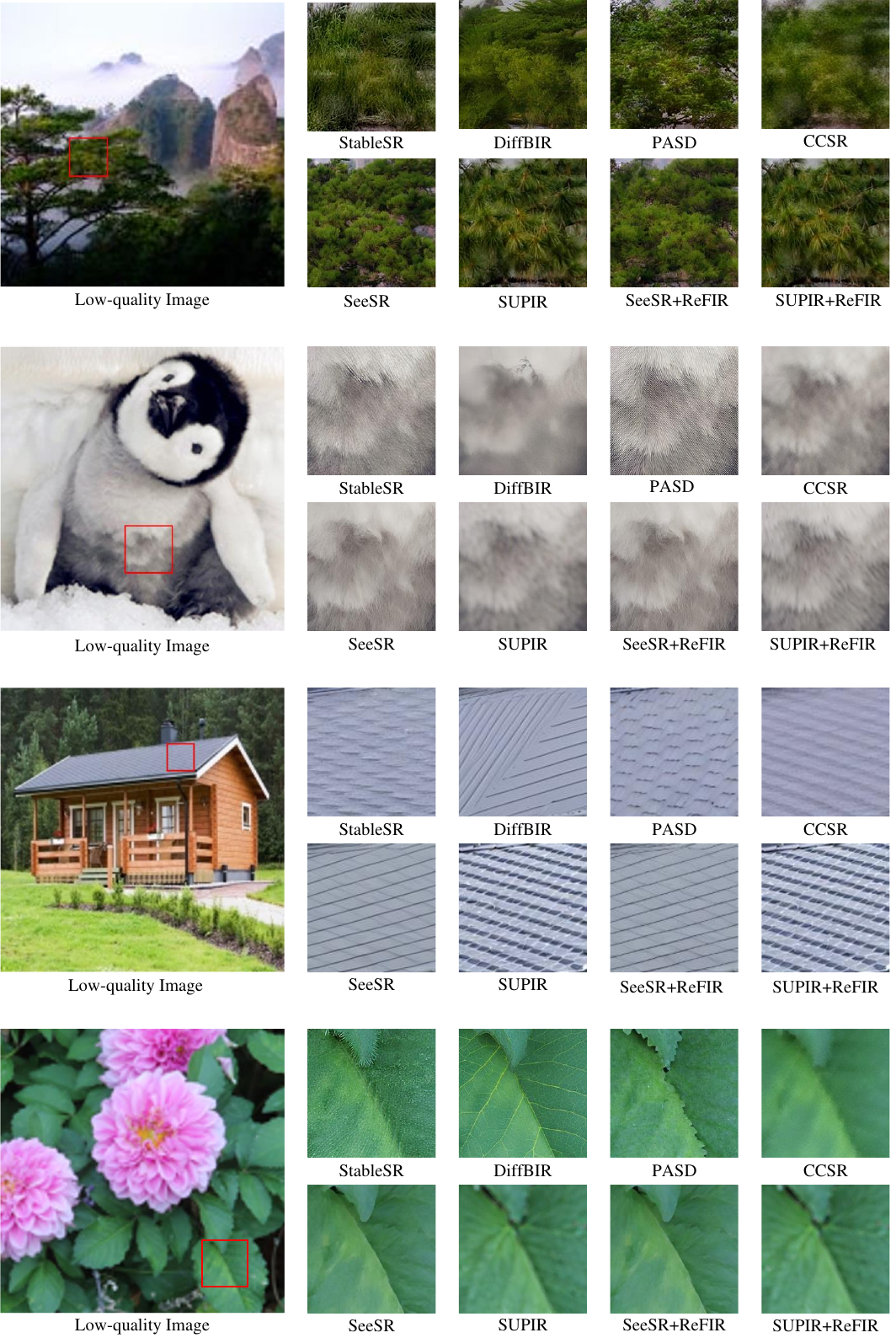}
 \caption{Additional qualitative comparison with state-of-the-art methods on RealPhoto60~\cite{yu2024supir}. Please zoom in for better effects.}
 \label{fig:supp-realPhoto}
 \vspace{-1mm}
\end{figure*}

\begin{figure*}[p]
\centering
 \includegraphics[width=1\linewidth]{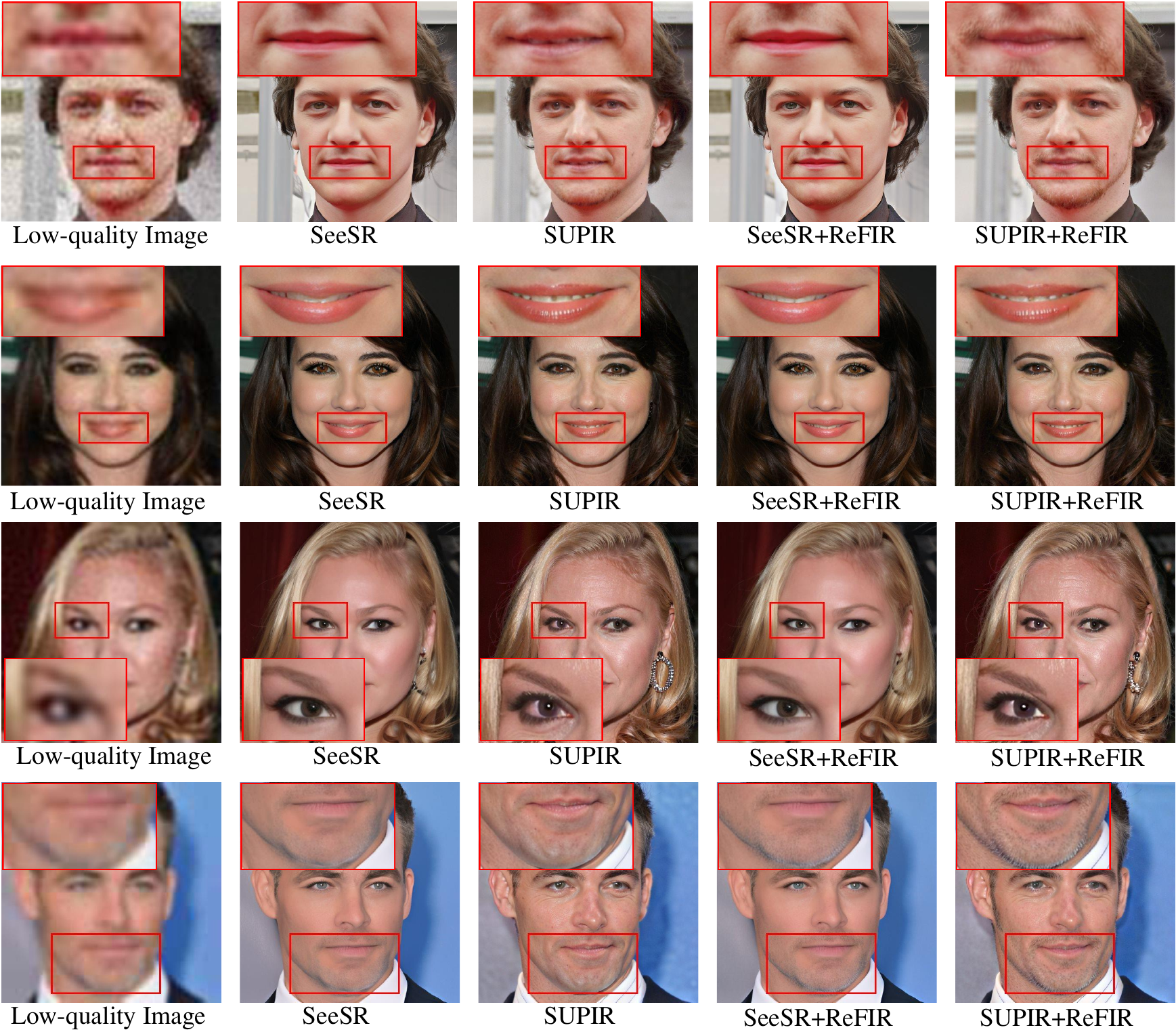}
 \caption{Visualization results of applying the proposed ReFIR to the downstream specific domain of blind face image restoration. Please zoom in for better effects.}
 \label{fig:supp-faceRestore}
 \vspace{-1mm}
\end{figure*}

In this section, we provide more visual results, which are organized as follows:

\noindent \begin{itemize}
     \item In \cref{fig:supp-pca}, we give more samples of the PCA visualization on the top three principal components of the self-attention layer latent.
    \item In \cref{fig:supp-cross-attn}, we give a visualization of the attention map from the cross-image injection, to help better understand the feasibility of cross-image attention. 
    \item \cref{fig:supp-realPhoto} gives more quantitative comparison results against the state-of-the-art method on real-world degradation without ground truth.
     \item \cref{fig:supp-faceRestore} gives the visualization results of the extension experiments of applying the proposed ReFIR to blind face image restoration.
\end{itemize}

\clearpage

\clearpage

\clearpage

\clearpage
\end{document}